\documentclass{article}

\usepackage{arxiv}

\usepackage[utf8]{inputenc}
\usepackage{amsmath}
\usepackage[figuresleft]{rotating}
\usepackage{txfonts}%
\usepackage{multirow}
\usepackage{array}
\usepackage{xcolor}
\usepackage[hidelinks]{hyperref}
\usepackage{algorithm}
\usepackage{booktabs} 
\usepackage{subcaption}
\usepackage{enumitem}
\usepackage[noend]{algpseudocode}

\DeclareMathOperator*{\argmax}{arg\,max}

\newcolumntype{L}[1]{>{\raggedright\let\newline\\\arraybackslash\hspace{0pt}}m{#1}}
\newcolumntype{C}[1]{>{\centering\let\newline\\\arraybackslash\hspace{0pt}}m{#1}}
\newcolumntype{R}[1]{>{\raggedleft\let\newline\\\arraybackslash\hspace{0pt}}m{#1}}

\definecolor{myML}{HTML}{08519C}
\definecolor{myOR}{HTML}{3182BD}
\definecolor{myNET}{HTML}{6BAED6}
\definecolor{myPS}{HTML}{BDD7E7}
\definecolor{mySTAT}{HTML}{EFF3FF}

\title{SoftED: Metrics for Soft Evaluation of Time Series Event Detection}

\author{
	Rebecca~Salles \\
	CEFET/RJ\\
	\texttt{rebeccapsalles@acm.org} \\
	\And
	Janio~Lima \\
	CEFET/RJ\\
	\texttt{janio.lima@eic.cefet-rj.br} \\
	\AND
	Michel~Reis\\
	CEFET/RJ\\
	\texttt{michel.reis@eic.cefet-rj.br} \\
	\And
	Rafaelli~Coutinho \\
	CEFET/RJ\\
	\texttt{rafaelli.coutinho@cefet-rj.br} \\
	\AND
	Esther~Pacitti \\
	University of Montpellier \& INRIA\\
	\texttt{esther.Pacitti@lirmm.fr} \\
	\And
	Florent~Masseglia \\
	University of Montpellier \& INRIA\\
	\texttt{florent.masseglia@inria.fr} \\
	\And
	Reza~Akbarinia \\
	University of Montpellier \& INRIA\\
	\texttt{reza.akbarinia@inria.fr} \\
	\AND
	Chao~Chen \\
	University of Nottingham\\
	\texttt{chao.chen@nottingham.ac.uk} \\
	\And
	Jonathan Garibaldi \\
	University of Nottingham\\
	\texttt{jon.garibaldi@nottingham.ac.uk} \\
	\AND
	Fabio Porto \\
	LNCC\\
	\texttt{fporto@lncc.br} \\
	\And
	Eduardo~Ogasawara\\
	CEFET/RJ\\
	\texttt{eogasawara@ieee.org} \\
}

\begin{document}
	\maketitle

	\begin{abstract}
Time series event detectors are evaluated mainly by standard classification metrics focusing solely on detection accuracy. However, inaccuracy in detecting an event can often result from its preceding or delayed effects reflected in neighboring detections. These detections are valuable to trigger necessary actions or help mitigate unwelcome consequences. In this context, current metrics are insufficient and inadequate for the context of event detection. There is a demand for metrics that incorporate both the concept of time and temporal tolerance for neighboring detections. Inspired by fuzzy sets, this paper introduces SoftED metrics, a new set designed for soft evaluating event detectors. They enable the evaluation of the detection accuracy and the degree to which their detections represent events. A new general protocol inspired by competency questions is also introduced to evaluate temporal tolerant metrics for event detection. The SoftED metrics can improve event detection evaluations by associating events and their representative detections, incorporating temporal tolerance in over 36\% of the overall conducted detector evaluations compared to the usual classification metrics. Following the proposed evaluation protocol, SoftED metrics were evaluated by domain specialists who indicated their contribution to detection evaluation and method selection.

Please cite the updated journal paper published at \url{https://doi.org/10.1016/j.cie.2024.110728}.

	\end{abstract}
	
\section{Introduction} \label{Intro}

In time series analysis, it is often possible to observe a significant change in observations at a certain instant or interval. Such a change generally characterizes the occurrence of an event \cite{guralnik_event_1999}. An event can represent a phenomenon with a defined meaning in a domain. Event detection is the process of identifying events in a time series. With this process, we may be interested in learning/identifying past events \cite{pimentel_review_2014, ding_experimental_2014,ahmed_survey_2016, alevizos_probabilistic_2017,sebestyen_taxonomy_2018,zhou_nonparametric_2019,pang_deep_2021,wang_progress_2019}, identifying events in real-time (online detection) \cite{zhang_outlier_2010,ahmad_unsupervised_2017,ariyaluran_habeeb_real-time_2019,munir_comparative_2019}, or even predicting future events before they happen (event prediction) \cite{yan_research_2004,salfner_survey_2010, molaei_analytical_2015, gmati_taxonomy_2019, zhao_event_2021}. It is recognized as a basic function in surveillance and monitoring systems and has gained much attention in research for application domains involving large datasets from critical systems \cite{pimentel_review_2014}.

To address the task of time series event detection, several methods have been developed and are surveyed in the literature \cite{hodge_survey_2004,chandola_anomaly_2009,gupta_outlier_2014,choudhary_runtime-efficacy_2017,cook_anomaly_2020,braei_anomaly_2020,blazquez-garcia_review_2021,schmidl_anomaly_2022}. Each detection method (detector, for short) specializes in time series that present different characteristics or make assumptions about the data distribution. Therefore, assessing detection performance is important to infer their adequacy for a particular application \cite{fontugne_mawilab_2010}. In this case, detection performance refers to how accurate an event detector is at identifying events in a time series. Detection performance is generally measured by classification metrics \cite{han_data_2022}.

Currently, standard classification metrics (around since the 1950s), including Recall, Precision, and F1, are usually adopted \cite{lavin_evaluating_2015,tatbul_precision_2018}. Although Accuracy is a specific metric \cite{han_data_2022}, the expression detection accuracy is henceforth used to refer to the ability of a method to detect events correctly. Classification metrics focus mainly on an analysis of detection accuracy. On the other hand, inaccuracy in event detection does not always indicate a bad result, especially when detections are sufficiently close to events.

\subsection{Motivating example and problem definition} \label{motivating_example}

\begin{figure}[!t]
\centering
\includegraphics[width=1\linewidth]{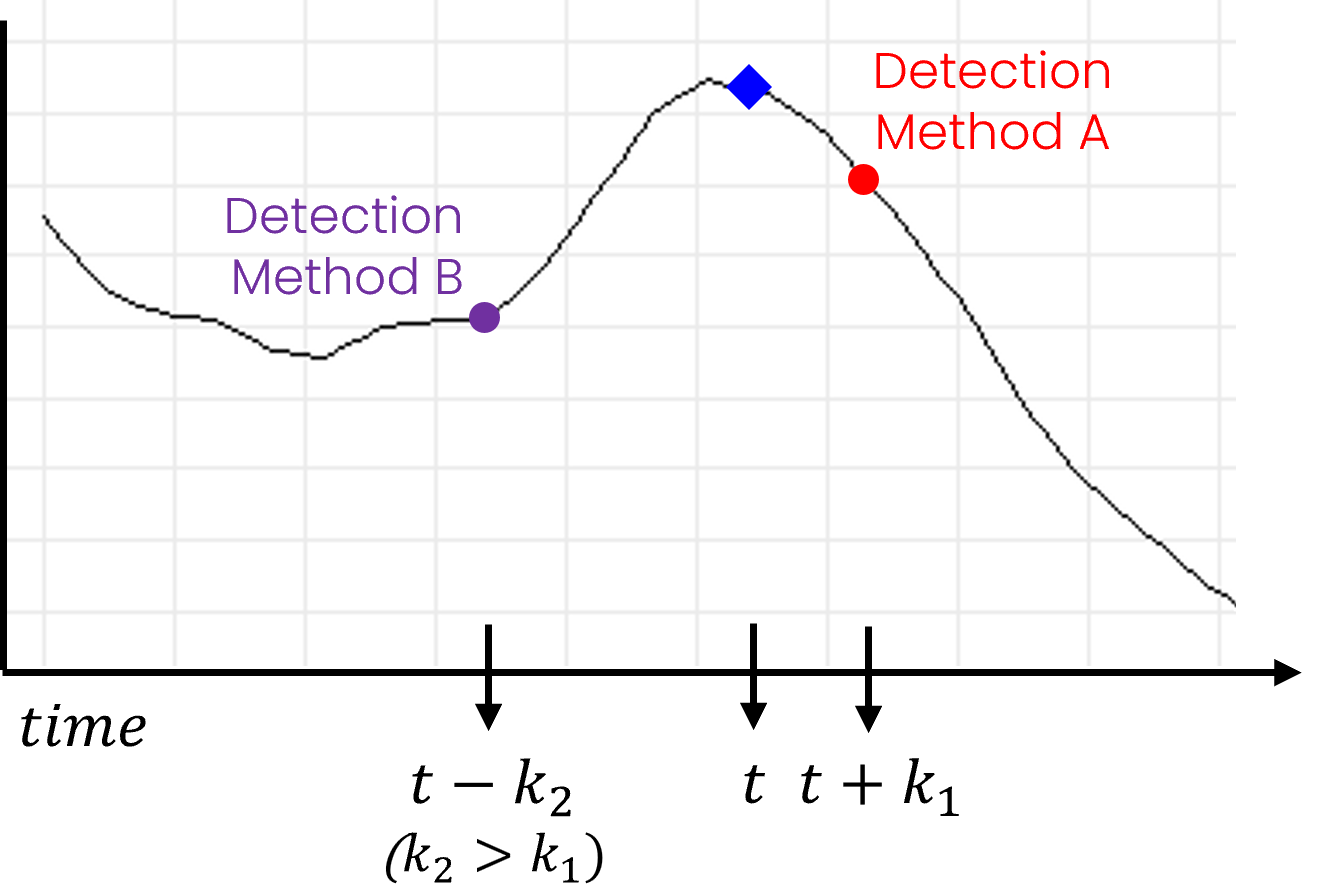}
\caption[Motivating example]{Example regarding the problem of evaluating the detection of an event at time $t$. Detector A detects an event at time $t+k_1$, while Detector B detects an event at time $t-k_2$ ($k_2>k_1$).}
\label{fig_motivating_example}
\end{figure}

This section gives an example of the problem of evaluating inaccurate event detections and defines the problem in the paper concerning the demand for adequate detection performance metrics. Consider, for example, a time series $X$ containing an event at time $t$, represented in Figure~\ref{fig_motivating_example}. Since detectors A and B are applied to $X$, a user must select one of them as the most adequate for the underlying application. Detector A detects an event at time $t+k_1$, while Detector B detects an event at time $t-k_2$ ($k_2>k_1$). As none of the methods could correctly detect the event at time $t$, based on the usual detection accuracy evaluation, the user would deem both inaccurate and disposable. 

However, inaccuracy in detecting an event can often result from its preceding or lingering effects. Take the adoption of a new policy in a business. While a domain specialist may consider the moment of policy enforcement as a company event, its effects on profit may only be detectable a few months later. On the other hand, preparations for policy adoption may be detectable in the antecedent months. Moreover, when accurate detections are not achievable, which is common, detection applications demand events to be identified as soon as possible \cite{lavin_evaluating_2015}, or early enough to allow necessary actions to be taken, mitigating possible critical system failures or help mitigate urban problems resulting from extreme weather events, for example. In this context, the results of Methods A and B would be valuable to the user. Note that while Detector B seems to anticipate the event, its application to $X$ and resulting detection is made after the event's occurrence. On the other hand, the detection of Detector A came temporally closer to the event, possibly more representative of its effects.

In this context, evaluating event detection is particularly challenging, and the detection accuracy metrics usually adopted are insufficient and inadequate for the task \cite{singh_demystifying_2017}. Standard classification metrics do not consider the concept of time, which is fundamental in the context of time series analysis, and do not reward early detection \cite{ahmad_unsupervised_2017}, for example, or any relevant neighboring detections. For the remainder of this paper, neighboring or close detections refer to detections whose temporal distance to events is within a desired threshold. Current metrics only reward true positives (exact matches in event detection). All other results are ``harshly'' and equally discredited.

In this case, there is a demand to soften the usual concept of detection accuracy and evaluate the methods while considering neighboring detections. However, state-of-the-art metrics designed for scoring anomaly detection, such as the NAB score \cite{lavin_evaluating_2015}, are still limited \cite{singh_demystifying_2017}, while also being biased towards results preceding events. To the best of our knowledge, no metrics available in the literature consider the concept of time and tolerance for detections sufficiently close to time series events. This paper focuses on addressing this demand. 

\subsection{Contribution} \label{contribution}
This paper introduces the SoftED metrics, a new set of metrics for evaluating event detectors regarding their detection accuracy and ability to produce neighboring detections to events. Inspired by fuzzy sets, SoftED metrics assess the degree to which a detection represents a particular event. It is an innovative approach since, while fuzzification is commonly applied to time series observations \cite{hullermeier_fuzzy_2011}, in SoftED, the fuzzification occurs in the time dimension \cite{zadeh_fuzzy_1965}. Hence, they incorporated time and temporal tolerance in evaluating the accuracy of event detection. These are the scenarios that domain specialists and users often face. Up until now, there have been no standard or adequate evaluation metrics for event detection in such scenarios. SoftED metrics soften the standard classification metrics, which are considered in this paper as hard metrics, to support the decision-making regarding the most appropriate method for a given application.

This paper also contributes by introducing a new general protocol for evaluating the performance of metrics in time series event detection. This protocol is designed to evaluate metrics that incorporate temporal tolerance into detection evaluation. It is inspired by a strategy commonly adopted in ontology design. It is based on using competency questions \cite{uschold_ontologies_1996,noy_state_1997}.

Computational experiments were conducted to analyze the contribution of the developed metrics against the usual hard and state-of-the-art metrics \cite{lavin_evaluating_2015}. The results illustrate that SoftED can improve event detection evaluation by associating events and their representative (neighboring) detections, incorporating temporal tolerance in over 36\% of the conducted experiments compared to usual hard metrics. More importantly, following the proposed evaluation protocol, we evaluated the SoftED metrics based on the devised competency questions. Domain specialists confirmed the contribution of SoftED metrics to the problem of detector evaluation.

The remainder of this paper is organized as follows. Section~\ref{background} provides concepts on time series event detection and reviews the literature on detection performance metrics for detection evaluation. Section~\ref{softMetrics} formalizes the developed SoftED metrics. Section~\ref{experiment} presents a quantitative and qualitative experimental evaluation of the developed metrics and their empirical results. Finally, conclusions are made in Section~\ref{conclusion}.

\section{Literature review} \label{background}

This section provides relevant concepts on time series events and their detection and reviews the literature on detection performance metrics and related works. Events are pervasive in real-world time series, especially in the presence of nonstationarity \cite{guralnik_event_1999,salles_nonstationary_2019}. Commonly, the occurrence of an event can be detected by observing anomalies or change points. Most event detectors in the literature specialize in identifying a specific type of event. There exist methods that can detect multiple events in time series, generally involving the detection of both anomalies and change points \cite{lawhern_detect:_2013, aminikhanghahi_survey_2017, zhao_event_2021}. Nonetheless, these methods are still scarce. This paper approaches methods for detecting anomalies, change points, or both.

\subsection{Time series events} \label{events}

Events correspond to a phenomenon, generally pre-defined in a particular domain, with an inconstant or irregular occurrence relevant to an application. In time series, events represent significant changes in expected behavior at a certain time or interval \cite{guralnik_event_1999}. In general, instant events of a given time series $X$~=~\textless $x_1, x_2, x_3, \cdots, x_n$\textgreater, can be identified in a simplified way by $e(X,k,\sigma)$ using the Equation~\ref{eq_events}, where $k$ represents the length of nearby observations\footnote{Due to limited space, the general formalization of event intervals lie outside the scope of this paper.}.
A temporal component (TC) for an observation $x_t$ is expressed as $tc(x_t)$. The TC can refer to the observation itself or its instant trend, respectively represented as $x_t$ and $tr(x_t)$.

Considering a typical autoregressive behavior \cite{gujarati_essentials_2021}, one can expect a TC for $x_t$ to be related to previous observations. Let $ep$ be the expected TC for $x_t$ based on the previous $k$ observations, where $ep(tc(x_t), k) = E(tc(x_t) ~|~ tc(x_{t-k}),\dots,tc(x_{t-1}))$. Analogously, one can also expect TC for $x_t$ to be explained from the following observations \cite{lima_forward_2022}. So let $ef$ be the expected TC for $x_t$ based on the following $k$ observations, where $ef(tc(x_t), k) = E(tc(x_t) ~|~ tc(x_{t+1}), \dots, tc(x_{t+k}))$.
If a TC for $x_t$ escapes the expected value above a threshold $\sigma$ based on previous or following $k$ observations \cite{gujarati_essentials_2021,lima_forward_2022}, it can be considered an event. Equation \ref{eq_events} also considers an event if the expected TC from previous and following $k$ observations are different above the threshold $\sigma$.

\begin{equation}
\begin{split}
e(X, k, \sigma) = \{t, |tc(x_t) - ep(tc(x_t), k)| > \sigma \\ \vee ~ |tc(x_t) - ef(tc(x_t), k)| > \sigma \\ \vee ~ |ep(tc(x_t), k) - ef(tc(x_t), k)| > \sigma \}
\end{split}
\label{eq_events}
\end{equation}

\paragraph{Anomalies}

Most commonly, events detected in time series refer to anomalies. Anomalies appear not to be generated by the same process as most of the observations in the time series \cite{chandola_anomaly_2009}. Thus, anomalies can be modeled as isolated observations of nearby data. In this case, an event identified in $x_t$ can be considered an anomaly if it escapes expected TC before and after time point $t$ according to $a(X,k,\sigma)$ in Equation \ref{eq_anomalies}. Generally, anomalies are identified by deviations from the time series inherent trend. 

\begin{equation}\label{eq_anomalies}
\begin{split}
a(X,k,\sigma) = \{t, |tc(x_t) - ep(tc(x_t), k)| > \sigma \\ \wedge ~ |tc(x_t) - ef(tc(x_t), k)| > \sigma\}
\end{split}
\end{equation}

\paragraph{Change points}

Change points in a time series are the points or intervals in time that represent a transition between different states in a process that generates the time series \cite{takeuchi_unifying_2006}. In this case, a change point event identified in time $t$ follows the expected behavior observed before or after the time point $t$, but not both at the same time according to $cp(X,k,\sigma)$ in Equation \ref{eq_change_points}. It can also refer to a significant difference between the expected trend before or after time point $t$.

\begin{equation}\label{eq_change_points}
\begin{split}
cp(X, k, \sigma) = \{t, ~ (|tc(x_t) - ep(tc(x_t), k)| > \sigma \\ \veebar ~ |tc(x_t) - ef(tc(x_t), k)| > \sigma) \\ \vee ~ (|ep(tr(x_t), k) - ef(tr(x_t), k)| > \sigma) \}
\end{split}
\end{equation}

\subsection{Event detection} \label{eventDetection}

Event detection is the process of identifying the occurrence of such events based on data analysis. It is recognized as a basic function in surveillance and monitoring systems. Moreover, it becomes even more relevant for applications based on time series and sensor data analysis \cite{pimentel_review_2014}. Event detectors found in the literature are usually based on model deviation analysis, classification-based analysis, clustering-based analysis, domain-based analysis, or statistical techniques \cite{chandola_anomaly_2009,han_data_2022,pimentel_review_2014,alevizos_probabilistic_2017}. Regardless of the adopted detection strategy, an important aspect of any event detector is how the events are reported. Typically, the outputs produced by event detectors are either scores or labels. Scoring detectors assign an anomaly score to each instance in the data depending on the degree to which that instance is considered an anomaly. On the other hand, labeling detectors assign a label (normal or anomalous) to each data instance. Such methods are the most commonly found in the literature \cite{chandola_anomaly_2009}. 

\subsection{Detection performance metrics} \label{usualMerics}

Detectors might vary performance under different time series \cite{fanaee-t_event_2014}. Therefore, there is a demand to compare the results they provide. Such a process aims to guide the choice of suitable methods for detecting events of a time series in a particular application. For comparing event detectors, standard classification metrics, such as F1, Precision, and Recall, are usually adopted \cite{han_data_2022,lavin_evaluating_2015,tatbul_precision_2018}.

As usual, the standard classification metrics depend on measures of true positives ($TP$), true negatives ($TN$), false positives ($FP$), and false negatives ($FN$) \cite{han_data_2022}. In event detection, the $TP$ refers to the number of events correctly detected (labeled) by the method. Analogously, $TN$ is the number of observations that are correctly not detected. On the other hand, the measure $FP$ is the number of detections that did not match any event, that is, false alarms. Analogously, $FN$ is the number of undetected events. Among the standard classification metrics, Precision and Recall are widely adopted. Precision reflects the percentage of detections corresponding to time series events (exactness), whereas Recall reflects the percentage of correctly detected events (completeness). Precision and Recall are combined in the F$_\beta$ metrics \cite{han_data_2022}. The F1 metric is also widely used to help gauge the quality of event detection balancing Precision and Recall \cite{tatbul_precision_2018}.

As discussed in Section \ref{motivating_example}, event detection is particularly challenging to evaluate, event detection is particularly challenging to evaluate. In this context, the Numenta Anomaly Benchmark (NAB) provided a common scoring algorithm for evaluating and comparing the efficacy of anomaly detectors \cite{lavin_evaluating_2015}. The NAB score metric is computed based on anomaly windows of observations centered around each event in a time series. Given an anomaly window, NAB uses the sigmoidal scoring function to compute the weights of each anomaly detection. It rewards earlier detections within a given window and penalizes $FP$s. Also, NAB allows the definition of application profiles: standard, reward low $FP$s, and reward low $FN$s. Based on the window size, the standard profile gives relative weights to $TP$s, $FP$s, and $FN$s.

Nonetheless, the NAB scoring system presents challenges for its usage in real-world applications. For example, the anomaly window size is automatically defined as 10\% of the time series size, divided by the number of events it contains, and values generally not known in advance, especially in streaming environments. Furthermore, Singh and Olinsky \cite{singh_demystifying_2017} pointed out the scoring equations' poor definitions and arbitrary constants. Finally, score values increase with the number of events and detections. Every user can tweak the weights in application profiles, making it difficult to interpret and benchmark results from other users or setups.

In addition to NAB \cite{lavin_evaluating_2015}, this section presents other works related to the problem of analyzing and comparing event detection performance \cite{chandola_anomaly_2009}. Recent works focus on the development of benchmarks to evaluate univariate time series anomaly detectors \cite{jacob_exathlon_2021, boniol_theseus_2022, wenig_phillipwenighpide_timeeval_2022}. Jacob et al. \cite{jacob_exathlon_2021} provide a comprehensive benchmark for explainable anomaly detection over high-dimensional time series. In contrast, the benchmark developed by Boniol et al. \cite{boniol_theseus_2022} allows the user to assess the advantages and limitations of anomaly detectors and detection accuracy metrics.

Standard classification metrics are generally used for evaluating the ability of an algorithm to distinguish normal from abnormal data samples \cite{chandola_anomaly_2009, tatbul_precision_2018}. Aminikhanghahi and Cook \cite{aminikhanghahi_survey_2017} review traditional metrics for change point detection evaluation, such as Sensitivity, G-mean, F-Measure, ROC, PR-Curve, and MSE. Detection evaluation measures have also been investigated in the areas of sequence data anomaly detection \cite{chandola_comparative_2008}, time series mining and representation \cite{ding_querying_2008}, and sensor-based human activity learning \cite{cook_activity_2015, ward_performance_2011}.

Metrics found in the literature are mainly designed to evaluate the detection of instant anomalies. However, many real-world event occurrences extend over an interval (range-based). Motivated by this, Tatbul et al. \cite{tatbul_precision_2018} and Paparrizos et al. \cite{paparrizos_volume_2022} extend the well-known Precision and Recall metrics, and the AUC-based metrics, respectively, to measure the accuracy of detection algorithms over range-based anomalies. Other recent metrics developed for detecting range-based time series anomalies are also included in the benchmark of Boniol et al. \cite{boniol_theseus_2022}. In addition, Wenig et al. \cite{wenig_phillipwenighpide_timeeval_2022} published a benchmarking toolkit for algorithms designed for detecting anomalous subsequences in time series \cite{boniol_series2graph_2020, boniol_sand_2021}.

Few works opt to evaluate event detection algorithms based on metrics other than traditional ones. For example, Wang, Vuran, and Goddard \cite{wang_analysis_2011} calculate the delay until an individual node and the delivery delay in a transmission network detect an event. The work presents a framework for capturing delays in detecting events in large-scale WSN networks with a time-space simulation. Conversely, Tatbul et al. \cite{tatbul_precision_2018} also observe the neighborhood of event detections, not to calculate detection delays, but to evaluate positional tendency in anomaly ranges. Our previous work uses the delay measure to evaluate the bias of algorithms to detect real events in time series \cite{escobar_evaluating_2021}. It furthers a qualitative analysis of the tendency of algorithms to detect before or after the occurrence of an event.

Under these circumstances, there is still a demand for event detection performance metrics that incorporate both the concept of time and tolerance for detections sufficiently close to time series events. Therefore, this paper contributes by introducing new metrics for evaluating methods regarding their detection accuracy and considering neighboring detections, incorporating temporal tolerance for inaccuracy in event detection.

\section{SoftED}\label{softMetrics}

This paper adopts a distance-based approach to develop novel metrics to evaluate the performance of methods for detecting events in time series. The inspiration for the proposed solution is found in soft (or approximate) computing. Soft computing is a collection of methodologies that exploit tolerance for inaccuracy, uncertainty, and partial truth to achieve tractability, robustness, and low solution cost \cite{tettamanzi_soft_2013}. In this context, the main proposed idea is to soften the hard metrics (standard classification metrics) to incorporate temporal tolerance or inaccuracy in event detection. Such metrics seek to support the decision-making of the most appropriate method for a given application with a basis not only on the usual analysis of the detection accuracy but also on the analysis of the ability of a method to produce detections that are close enough to actual time series events. Henceforth, the proposed approach is Soft Classification Metrics for Event Detection or SoftED. This section formalizes the SoftED metrics.

\begin{figure*}[t!]
	\centering
	\includegraphics[width=0.7\textwidth]{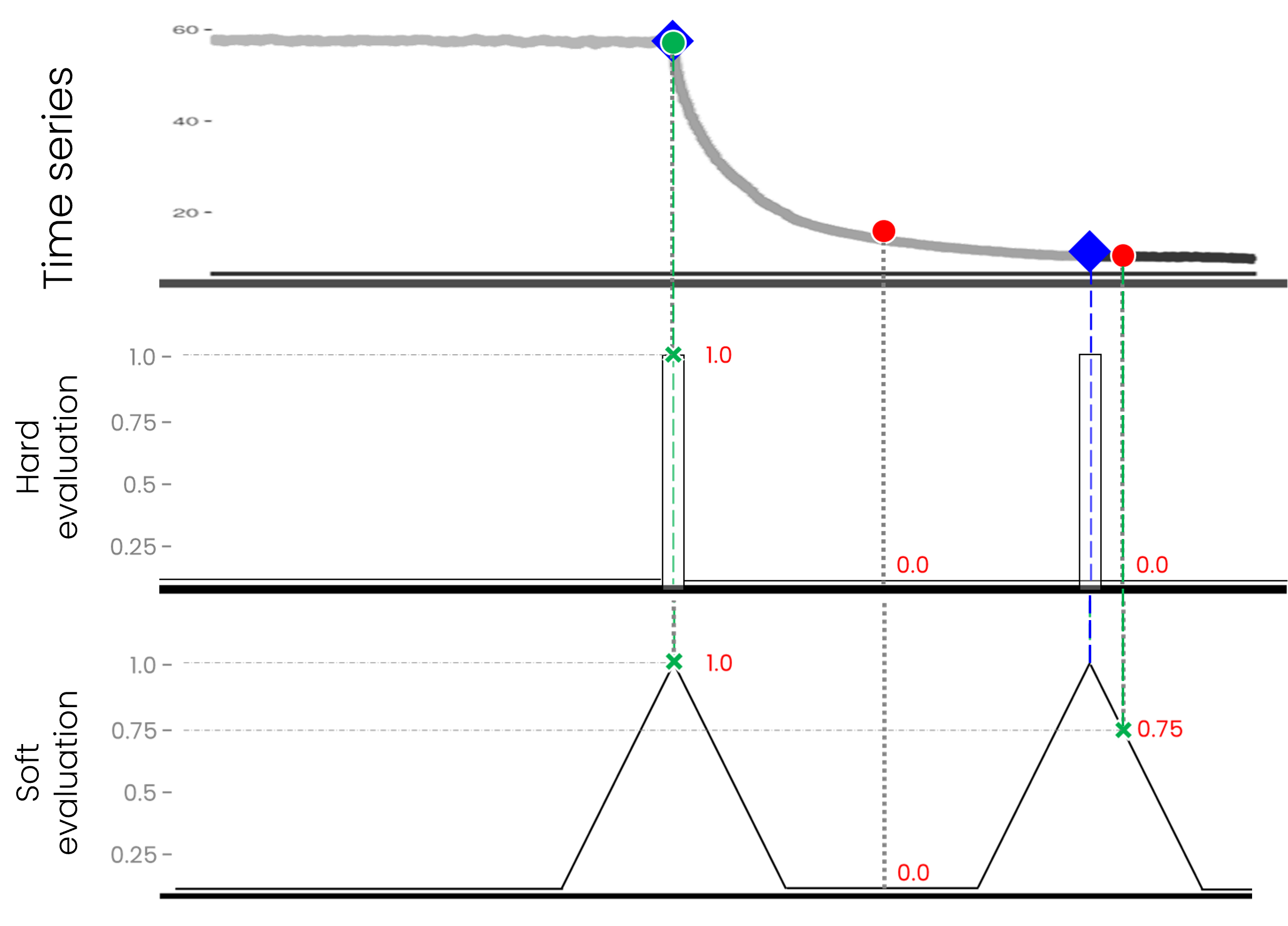}
	\caption{The general idea behind the proposed approach compares the standard ``hard'' evaluation and the ``soft'' evaluation of the event detection.}
 \label{fig_soft_evaluation}
\end{figure*}

Figure~\ref{fig_soft_evaluation} gives a general idea of the proposed approach, illustrating the key difference between the standard hard evaluation and the proposed soft evaluation. Blue rhombuses represent actual time-series events. Circles correspond to detections produced by a particular detector. The hard evaluation concerns a binary value regarding whether detection is a perfect match to the actual event. In this case, circles are green when they perfectly match the events and red when they do not. Conversely, soft evaluation assesses the degree to which detection relates to a particular event.

\subsection{Defining an event membership function} \label{SoftED_mu}

We incorporate a distance-based temporal tolerance for events. It is done by defining the relevance of a particular detection to an event. This section formalizes the proposed approach. Table~\ref{tbl_variable_def1} defines the main variables used in the formalization of SoftED. Given a time series $X$ of length $|X|$ containing a set of $m$ events, $E=\{e_1,e_2,\dots,e_m\}$, where $e_j$, $j=1,\dots,m$, is the j–th event in $E$ occurring at time point $t_{e_j}$. A particular detector applied to $X$ produces a set of $n$ detections, $D=\{d_1,d_2,\dots,d_n\}$, where $d_i$, $i=1,\dots,n$, is the i–th detection in $D$ indicating the time point $t_{d_i}$ as a detection occurrence.

\begin{table}[!ht]
\centering
\caption{Definition of main variables for the formalization of SoftED}
\label{tbl_variable_def1}
\setlength{\tabcolsep}{7.5pt}
\begin{tabular}{llp{4.6cm}}
\toprule
Var. & Value & Description \\ 
\midrule
$E$ & $\{e_1,e_2,\dots,e_m\}$ & set of time series events \\
$m$ & $|E|$ & number of events \\
$j$ & $1,\dots,m$ & event index \\
$e_j$ & $-$ & the j–th event in $E$ \\ 
$t_{e_j}$ & time point & time point where the $e_j$ occurs \\
\midrule
$D$ & $\{d_1,d_2,\dots,d_n\}$ & set of detections \\
$n$ & $|D|$ & number of detections \\
$i$ & $1,\dots,n$ & detection index \\
$d_i$ & $-$ & the i–th detection in $D$ \\
$t_{d_i}$ & time point & time point where $d_i$ occurs \\ 
\midrule
$k$ & time duration & constant of tolerance for event detections \\
\bottomrule
\end{tabular}
\end{table}

\begin{figure*}[!ht]
\centering
\begin{subfigure}{.49\textwidth}
 \centering
 \includegraphics[width=1\linewidth]{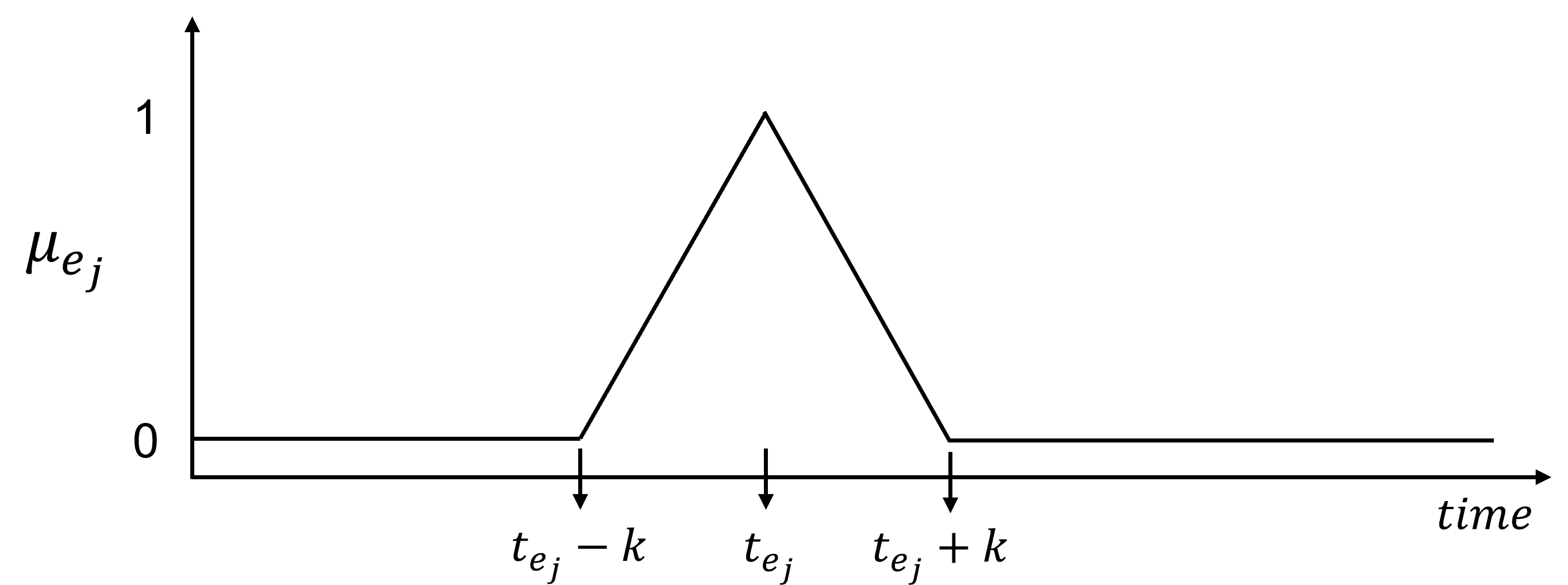}
 \caption{}
 \label{fig_mu_ej}
\end{subfigure}
\begin{subfigure}{.49\textwidth}
 \centering
 \includegraphics[width=1\linewidth]{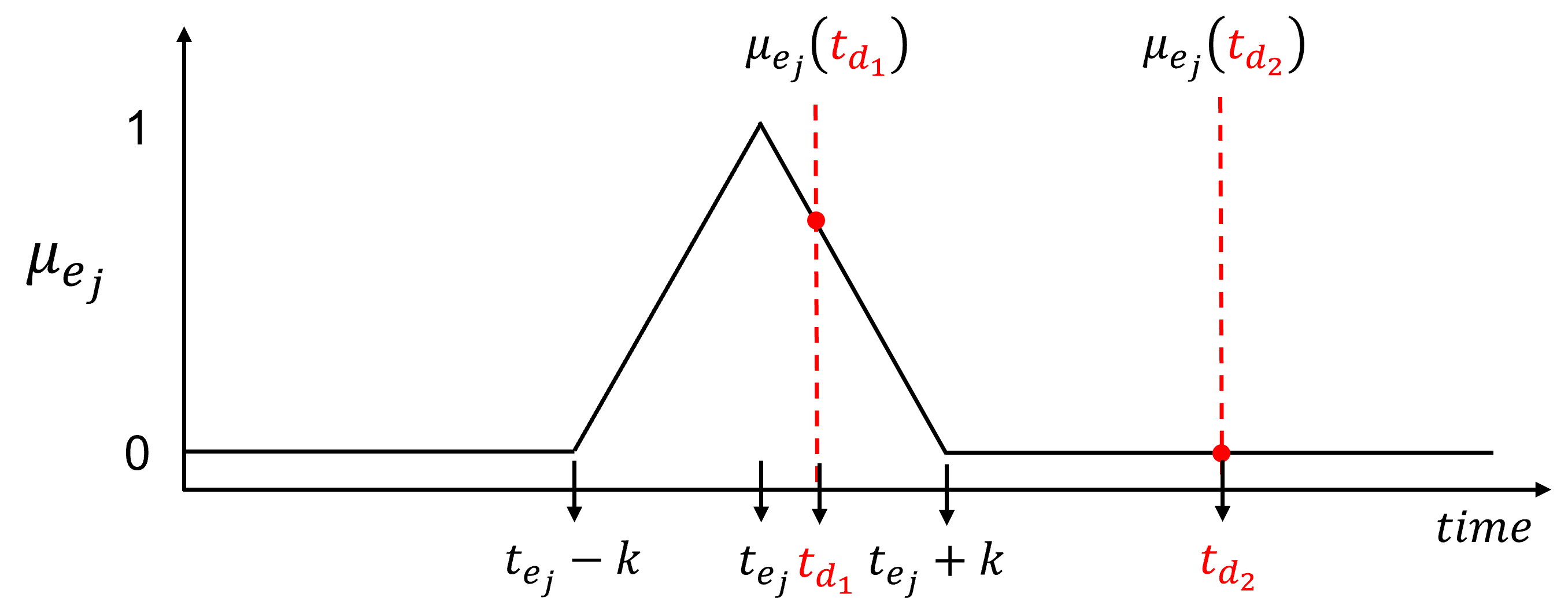}
 \caption{}
 \label{fig_mu_ej_t_di}
\end{subfigure}
\begin{subfigure}{.49\textwidth}
 \centering
 \includegraphics[width=1\linewidth]{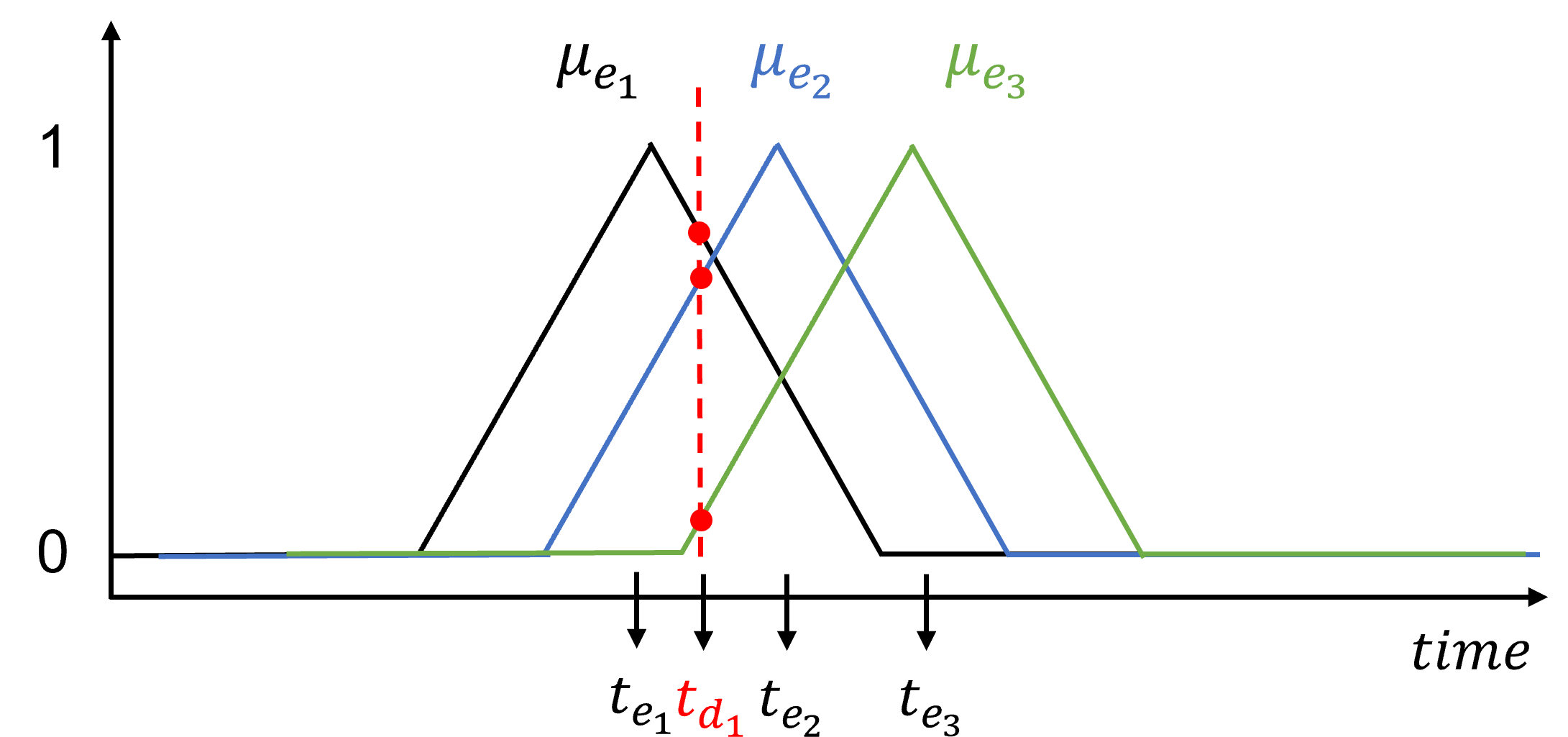}
 \caption{}
 \label{fig_scenario1}
\end{subfigure}
\begin{subfigure}{.49\textwidth}
 \centering
 \includegraphics[width=1\linewidth]{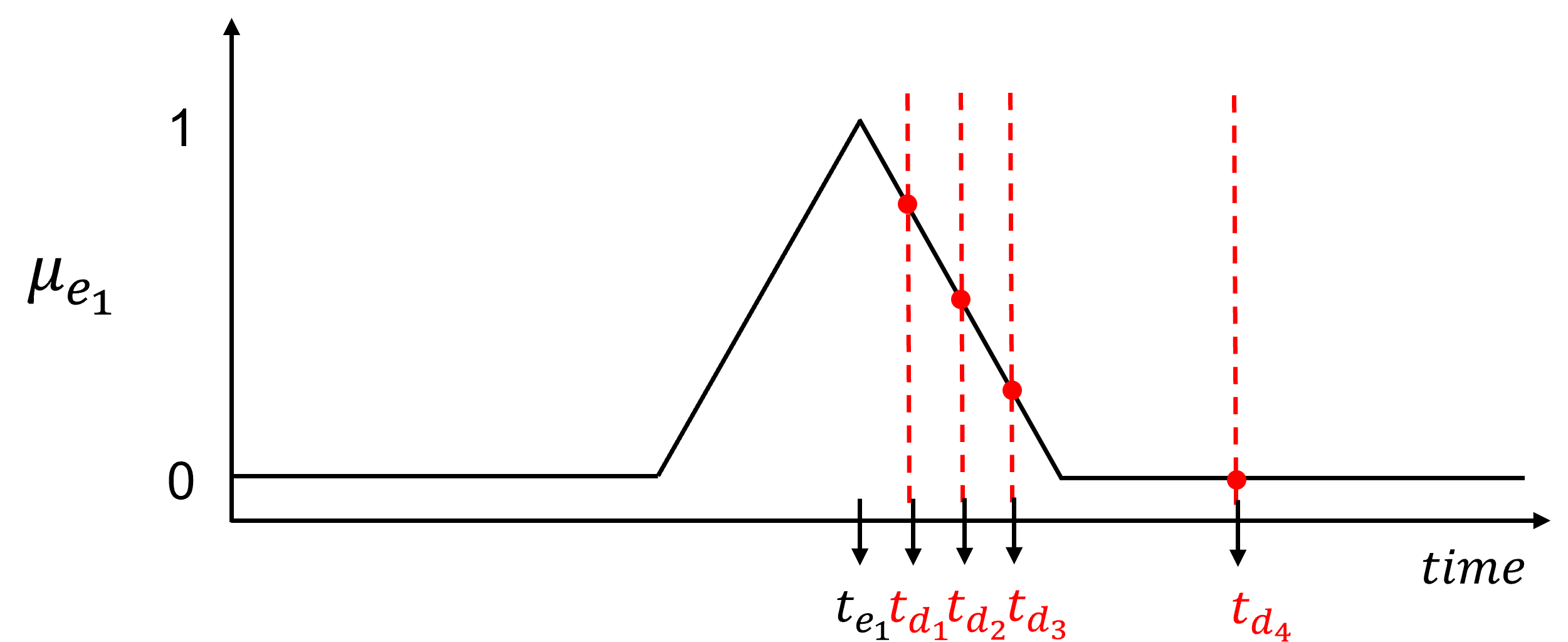}
 \caption{}
 \label{fig_scenario2}
\end{subfigure}
\caption[Auxiliary plots for comprehension of SoftED]{Auxiliary plots for comprehension of SoftED. (a) represents an event membership function $\mu_{e_j}(t)$. (b) represents $\mu_{e_j}(t)$ for detections $d_1$ and $d_2$. (c) depicts the example scenario containing one detection to many events, motivating the first constraint of SoftED. (d) depicts the example scenario containing many detections to a single event, motivating the second constraint of SoftED.}
\label{fig_barplots}
\end{figure*}

The degree to which a detection $d_i$ is relevant to a particular event $e_j$ is given by an event membership function $\mu_{e_j}(t)$ as defined in Equation~\ref{eq_mu_ej} and illustrated in Figure~\ref{fig_mu_ej}. The $\mu_{e_j}(t)$ represents a Euclidean distance function, which is of simple computation and interpretation. Moreover, this solution was inspired by Fuzzy sets, where we innovate by fuzzifying the time dimension rather than the time series observations \cite{zadeh_fuzzy_1965}. The definition of $\mu_{e_j}(t)$ considers the acceptable tolerance for inaccuracy in event detection for a particular domain application. The acceptable time range in which an event detection is relevant for allowing an adequate response reaction to a domain event is given by the constant $k$. For example, a meteorologist needs to provide alerts of extreme weather events at least 6 hours in advance, or an engineer needs to intervene in the operation of an overheated piece of machinery within 5 minutes before critical system failure. In this case, they might set $k$ to 6 hours or 5 minutes, respectively. However, this approach defines a domain-agnostic default value of $k$, set to $15$, defining a tolerance window of $30$ observations enough to hold the central limit theorem.

\begin{equation} \label{eq_mu_ej}
 \mu_{e_j}(t_{d_i}) = max\left(min\left(\frac{t_{d_i}-(t_{e_j}-k)}{k},\frac{(t_{e_j}+k)-t_{d_i}}{k}\right),0\right)
\end{equation}

Figure~\ref{fig_mu_ej_t_di} represents the evaluation of $\mu_{e_j}(t)$ for two detections, $d_1$ and $d_2$, produced by a particular detector. In this context, $\mu_{e_j}(t_{d_i})$ gives the extent to which a detection $d_i$ represents event $e_j$, or, in other words, its temporal closeness to a hard true positive (TP) regarding $e_j$. In that case, detection $d_1$ is closer to a TP, and $d_2$ lies outside the tolerance range given by $k$ and could be considered a false positive.

\subsection{Maintaining integrity with hard metrics} \label{SoftED_constraints}

We are interested in the SoftED metrics, which are still preserving concepts applicable to traditional (hard) metrics. In particular, the SoftED metrics are designed to express the same properties as their hard correspondents. Moreover, they are designed to maintain the reference to the perfect detection performance (score of $1$ as in hard metrics) and indicate how close a detector came to it. To achieve this goal, this approach defines constraints necessary for maintaining integrity concerning the standard hard metrics:
\begin{enumerate}
 \item A given detection $d_i$ must have only one associated score.
 \item The total score associated with a given event $e_j$ must not surpass $1$.
\end{enumerate}
 
The first constraint comes from the idea that the detection $d_i$ should not be rewarded more than once. It avoids the possibility of the total score for $d_i$ surpassing the perfect reference score of $1$. Take, for example, the first scenario, presented in Figure \ref{fig_scenario1}, in which we have one detection and many close events. The detection $d_1$ is evaluated for events $e_1$, $e_2$, and $e_3$ resulting in three different membership evaluation of $\mu_{e_1}(t_{d_1})$, $\mu_{e_2}(t_{d_1})$, and $\mu_{e_3}(t_{d_1})$, respectively. Nevertheless, to maintain integrity with hard metrics, a given detection $d_1$ must not have more than one score. Otherwise, $d_1$ would be rewarded three times, and its total score could surpass the score of a perfect match, which would be $1$.

To address this issue, we devise a strategy for attributing each detection $d_i$ to a particular event $e_j$. The adopted attribution strategy is based on the temporal distance between $d_i$ and $e_j$. It facilitates interpretation and avoids the need to solve an optimization problem for each detection. In that case, we attribute $d_i$ to the event $e_j$ that maximizes the membership evaluation $\mu_{e_j}(t_{d_i})$. This attribution is given by $E_{d_i}$ defined in Equation \ref{eq_E_di}. According to Figure \ref{fig_scenario1}, $d_1$ is attributed to event $e_1$ given the maximum membership evaluation of $\mu_{e_1}(t_{d_1})$. If there is a tie for the maximum membership evaluation of two or more events, $E_{d_i}$ represents the set of events to which $d_i$ is attributed. Consequently, we can also derive the set of detections attributed to each event $e_j$, $D_{e_j}$, defined by Equation \ref{eq_D_ej}. The addition of a detection $d_i$ to the set $D_{e_j}$ is further conditioned by the tolerance range, that is, a membership evaluation greater than $0$ ($\mu_{e_j}(t_{d_i})>0$).

\begin{equation} \label{eq_E_di} 
 E_{d_i} = \argmax_{e_j}\left(\mu_{e_j}(t_{d_i})\right)
\end{equation}
\begin{equation} \label{eq_D_ej} 
 D_{e_j} = \{d_i \mid E_{d_i}\supset e_j \land \mu_{e_j}(t_{d_i})>0\}
\end{equation}

The second constraint defined by this approach comes from the idea that a particular detector should not be rewarded more than once for detecting the same event $e_j$. It assures that the total score of detections for event $e_j$ does not surpass the perfect reference score of $1$. Take, for example, the second scenario, presented in Figure \ref{fig_scenario2}, in which many detections are attributed to the same event. The event $e_1$ is present in the sets $E_{d_1}$, $E_{d_2}$, $E_{d_3}$ and $E_{d_4}$. Moreover, $D_{e_1}$ contains $d_1$, $d_2$, and $d_3$. But to maintain integrity with hard metrics, the total score for $D_{e_1}$ ($\mu_{e_1}(t_{d_1})+\mu_{e_1}(t_{d_2})+\mu_{e_1}(t_{d_3})$) must not surpass the score of a perfect match.

To address this issue, we devise an analogous distance-based strategy for attributing a representative detection $d_i$ to each event $e_j$. In that case, we attribute to event $e_j$ the detection $d_i$, contained in $D_{e_j}$, that maximizes the membership evaluation $\mu_{e_j}(t_{d_i})$. This attribution is given by $\hat{d}_{e_j}$ defined in Equation \ref{eq_d'_ej}. According to Figure \ref{fig_scenario2}, $e_1$ is best represented by detection $d_1$ given the maximum membership evaluation of $\mu_{e_1}(t_{d_1})$. Consequently, we can compute the associated score for each event $e_j$ as $es(e_j)$, defined by Equation \ref{eq_Se}.

\begin{equation} \label{eq_d'_ej} 
 \hat{d}_{e_j} = \argmax_{d_i}\left(\{\mu_{e_j}(t_{d_i}) \mid d_i \in D_{e_j}\}\right)
\end{equation}

\begin{equation} \label{eq_Se} 
 es(e_j) = \mu_{e_j}(t_{\hat{d}_{e_j}})
\end{equation}

Finally, each detection $d_i$ produced by a particular detector is scored by $ds(d_i)$ defined in Equation \ref{eq_Sd}. Representative detections ($d_i=\hat{d}_{e_j}$) are scored based on $es(e_j)$. All other detections are scored $0$. This definition ensures the total score for detections of a particular method does not surpass the number of real events $m$ contained in the time series $X$. The Equation \ref{eq_sum_Sd} holds and maintains the reference to the score of a perfect detection Recall according to usual hard metrics. Furthermore, it penalizes false positives and multiple detections for the same event $e_j$. 

\begin{equation} \label{eq_Sd} 
 ds(d_i) = \begin{cases}
 es(e_j),& \text{if } \exists e_j \in E \mid d_i=\hat{d}_{e_j}\\
 0, & \text{otherwise}
\end{cases}
\end{equation}
\begin{equation} \label{eq_sum_Sd} 
 \sum_{i=1}^{n}ds(d_i) \leq m
\end{equation}

The pseudocode for calculating $ds(d_i)$ for each detection $d_i$ is given by Algorithm \ref{alg:soft_cores}, which defines the function $soft\_scores$. This function takes three arguments: $detections$, $events$, and $k$. The $detections$ argument is a logical vector of the same length as a time series $X$ ($|X|$), indicating the detections by a given method, and $events$ is a logical vector of the same length, indicating the true events in $X$. The $k$ argument is an optional parameter (default is set to $15$) that sets the temporal tolerance for a detection to be considered a match with a true event.

The function $soft\_scores$ in Algorithm \ref{alg:soft_cores} first takes the vectors $E$ and $D$ containing the indices of all events and detections in the input vectors $events$ and $detections$, respectively. Then, for each $i$-th detection, it finds the indices of all matching events within a distance of $k$ (line $7$) and stores them in $E_{match}$. If no matching events exist, $ds(d_i)$ is $0$. Otherwise, it calculates the membership scores between each matching event in $E_{match}$ and the $i$-th detection (line $11$) according to Equation~\ref{eq_mu_ej}. Finally, the soft score for each detection ($ds(d_i)$) is the maximum membership score among the matching events (line $12$), indicating the best match between the current detection and the events and respecting the constraints necessary for maintaining integrity with hard metrics defined by Equations \ref{eq_E_di} to \ref{eq_sum_Sd}. Lastly, a vector of soft scores for each detection ($ds$) is returned.

\begin{algorithm}[H] 
\caption{Computing soft scores for each detection} \label{alg:soft_cores}
\begin{algorithmic}[1]
\Function{soft\_scores}{$detections$, $events$, $k$}
 \State $E \gets \text{get\_indices}(events)$
 \State $D \gets \text{get\_indices}(detections)$
 \State $n \gets \text{length}(D)$
 \State $ds \gets \text{empty\_vector}(n)$
 \For{$i \gets 1$ to $n$}
 \State $E_{match} \gets \text{matching\_events}(D[i], E, k)$
 \If{$E_{match} \text{ is empty}$}
 \State $ds[i] \gets 0$
 \Else
 \State $scores \gets \text{membership\_scores}(D[i], E_{match}, k)$
 \State $ds[i] \gets \text{max}(scores)$
 \EndIf
 \EndFor
 \State \textbf{return} $ds$
\EndFunction

\end{algorithmic}
\end{algorithm}

\subsection{Computing the SoftED metrics} \label{SoftED_metrics}

The scores computed for each detection $d_i$, $ds(d_i)$, are used to create soft versions of the hard metrics TP, FP, TN, and FN, as formalized in Table \ref{tbl_soft_metrics}. In particular, while the value of TP gives the number of detections that perfectly matched an event (score of $1$), the sum of $ds(d_i)$ scores indicate the degree to which the detections of a method approximate the $m$ events contained in time series $X$ given the temporal tolerance of $k$ observations. Hence, the soft version of the TP metric, $\text{TP}_s$ is given by $\sum_{i=1}^{n}ds(d_i)$. Conversely, the soft version of FN, $\text{FN}_s$, indicates the degree to which a detector could not approximate the $m$ events in an acceptable time range. The $\text{FN}_s$ can then be defined as the difference between $\text{TP}_s$ and the perfect recall score ($m-\text{TP}_s$).

On the other hand, while the value of FP gives the number of detections that did not match an event (score of $0$), its soft version, $\text{FP}_s$, indicates how far the detections of a method came to the events contained in time series $X$ given the temporal tolerance of $k$ observations. In that sense, $\text{FP}_s$ is the complement of $\text{TP}_s$ and can be defined by $\sum_{i=1}^{n}\left(1-ds(d_i)\right)$. Finally, the soft version of TN, $\text{TN}_s$, indicates the degree to which a detector could avoid nonevent observations of $X$ ($|X|-m$). The $\text{TN}_s$ is given by the difference between $\text{FP}_s$ and the perfect specificity score ($(|X|-m)-\text{FP}_s$).

\begin{table}[!ht]
\centering
\caption{Formalization of SoftED Metrics}
\label{tbl_soft_metrics}
\setlength{\tabcolsep}{13.5pt}
\begin{tabular}{@{}rl rl@{}}
\toprule
$\text{TP}_s=$ & $\displaystyle\sum_{i=1}^{n}ds(d_i)$ & $\text{FN}_s=$ & $m-\text{TP}_s$ \\
\rule{0pt}{25pt}$\text{FP}_s=$ & $\displaystyle\sum_{i=1}^{n}\left(1-ds(d_i)\right)$ & $\text{TN}_s=$ & $(|X|-m)-\text{FP}_s$ \\ \bottomrule
\end{tabular}
\end{table}

Due to the imposed constraints described in Section \ref{SoftED_constraints}, the defined SoftED metrics $\text{TP}_s$, $\text{FP}_s$, $\text{TN}_s$, and $\text{FN}_s$, hold the same properties and the same scale as traditional hard metrics. Consequently, using the same characteristic formulas, they can derive soft versions of traditional scoring methods, such as Sensitivity, Specificity, Precision, Recall, and F1. Moreover, SoftED scoring methods still provide the same interpretation while including temporal tolerance for inaccuracy, which is pervasive in time series event detection applications. An implementation of SoftED metrics in R is made publicly available at Github\footnote{SoftED implementation, datasets and experiment codes: \url{https://github.com/cefet-rj-dal/softed}\label{footn:github_link}}.

\section{Experimental evaluation}\label{experiment}

SoftED metrics were submitted to an experimental evaluation to analyze their contribution against the traditional hard metrics and the NAB score, the current state-of-the-art detection scoring methods \cite{lavin_evaluating_2015}. The SoftED metrics are evaluated based on both complementary analyses: quantitative and qualitative. For that, a large set of computational experiments were performed with the application of several different methods for event detection in real-world and synthetic time series datasets containing ground truth event data. Detection results were evaluated based on SoftED, hard, and NAB metrics. First, this section describes the adopted time series datasets and experimental settings. Finally, the quantitative and qualitative results are presented.

\subsection{Datasets}
This section presents the datasets selected for evaluating the SoftED metrics. The selected datasets are widely available in the literature and are composed of simulated and real-world time series regarding several different domain applications such as water quality monitoring (GECCO) \cite{rehbach_gecco_2018}\footnote{The GECCO dataset is provided by the R-package EventDetectR \cite{moritz_eventdetectr_2020}.}, network service traffic (Yahoo) \cite{webscope_labeled_2015}, social media (NAB) \cite{ahmad_unsupervised_2017}, oil well exploration (3W) \cite{vargas_realistic_2019}, and public health (NMR) \footnote{The NMR dataset was produced by Fiocruz and comprised data on neonatal mortality in Brazilian health facilities from 2005 to 2017. It is publicly available at \url{https://doi.org/10.7303/syn23651701}.}, among others. The selected datasets present over $6$ hundred representative time series containing different events. In particular, GECCO, Yahoo, and NAB contain mostly anomalies, while 3W and NMR contain mostly change points. Moreover, the datasets present different nonstationarity and statistical properties to provide a more thorough discussion of the effects of the incorporated temporal tolerance on event detection evaluation on diverse datasets.

\subsection{Experimental settings}

For evaluating SoftED metrics, a set of up to $12$ different event detectors were applied to all time series in the adopted datasets, totalizing $4,026$ event detection experiments. Each experiment comprised an offline detection application, where the methods had access to the entire time series given as input. The applied detectors are implemented and publicly available in the \emph{Harbinger} framework \cite{salles_harbinger_2020}. It integrates and enables the benchmarking of different state-of-the-art event detectors. These methods encompass searching for anomalies and change points using statistical, volatility, proximity, and machine learning methods. The adopted methods are described in detail by Escobar et al. \cite{escobar_evaluating_2021}, namely: the Forward and Backward Inertial Anomaly Detector (FBIAD) \cite{lima_forward_2022}, K-Nearest Neighbors (KNN-CAD) \cite{gammerman_hedging_2007}, anomalize (based on time series decomposition \cite{dudek_neural_2016,shumway_time_2017}) \cite{gupta_outlier_2014,dancho_anomalize_2020}, and GARCH \cite{carmona_statistical_2013}, for anomaly detection; the Exponentially Weighted Moving Average (EWMA) \cite{raza_ewma_2015}, seminal method of detecting change points (SCP) \cite{guralnik_event_1999}, and ChangeFinder (CF) \cite{takeuchi_unifying_2006}, for change point detection; and the machine learning methods based on the use Feed-Forward Neural Network (NNET) \cite{riese_supervised_2020}, Convolutional Neural Networks (CNN) \cite{guo_simple_2017,lim_time-series_2021}, Support Vector Machine (SVM) \cite{chauhan_problem_2019,rahul_advanced_2021}, Extreme Learning Machine (ELM) \cite{ismaeel_using_2015,tang_extreme_2016}, and K-MEANS \cite{muniyandi_network_2012}, for general purpose event detection.

In each experiment, detectors were evaluated using the hard metrics Precision, Recall, and F1. Among them, the F1 was the main metric used for comparison. The NAB score was also computed with the standard application profile for each detector result. The NAB scoring algorithm is implemented and publicly available in the R-package \textit{otsad} \cite{iturria_otsad_2019}. 
Anomaly window sizes were automatically set. During the computation of the NAB score, confusion matrix metrics are built. Based on these metrics, the F1 metric of the NAB scoring approach was also computed. Finally, the SoftED metrics were computed for soft evaluation of the applied event detectors. In particular, this experimental evaluation sets the constant of temporal tolerance, $k$, to its default value ($15$) unless stated otherwise. Nevertheless, $k$ also experimented with values in $\{30,45,60\}$ for sensitivity analysis. 

The datasets and codes used in this experimental evaluation were available for reproducibility\footref{footn:github_link}.

\subsection{Quantitative analysis} \label{quanti}

This section presents the quantitative analysis of the SoftED metrics. The main goal of this analysis is to assess the effects of temporal tolerance incorporated by SoftED in event detection evaluation. In particular, this section intends to assess (i) whether the proposed metrics can incorporate temporal tolerance to event detection evaluation and, if so, (ii) whether the incorporated temporal tolerance affects the selection of detectors. Answering both questions demands an experimental evaluation to compare the proposed metrics against other baseline and state-of-the-art metrics.

\paragraph{Quantitative experiment 1} 
The first experiment assesses the number of times SoftED metrics considered more $TP$s while evaluating detectors to answer whether SoftED can incorporate temporal tolerance to event detection evaluation. We are interested in comparing SoftED F1 and hard F1 metrics for that. Time series detections where SoftED F1 was higher than its corresponding hard metric represent the incorporation of temporal tolerance. In contrast, detection results that maintained an unchanged F1 score had their evaluation confirmed, representing either perfect recall scenarios in which no tolerance is needed or scenarios with a low rate of neighboring detections in which there are few opportunities for tolerance. Finally, there are inaccurate results, with detections that did not allow temporal tolerance, presenting zero Precision/Recall, and no detections were sufficiently close to events given the defined tolerance level ($k=15$). In the latter case, the F1 metric cannot be computed.

\begin{figure}[!t]
\centering
\begin{subfigure}{.49\textwidth}
 \centering
 \includegraphics[width=1\linewidth]{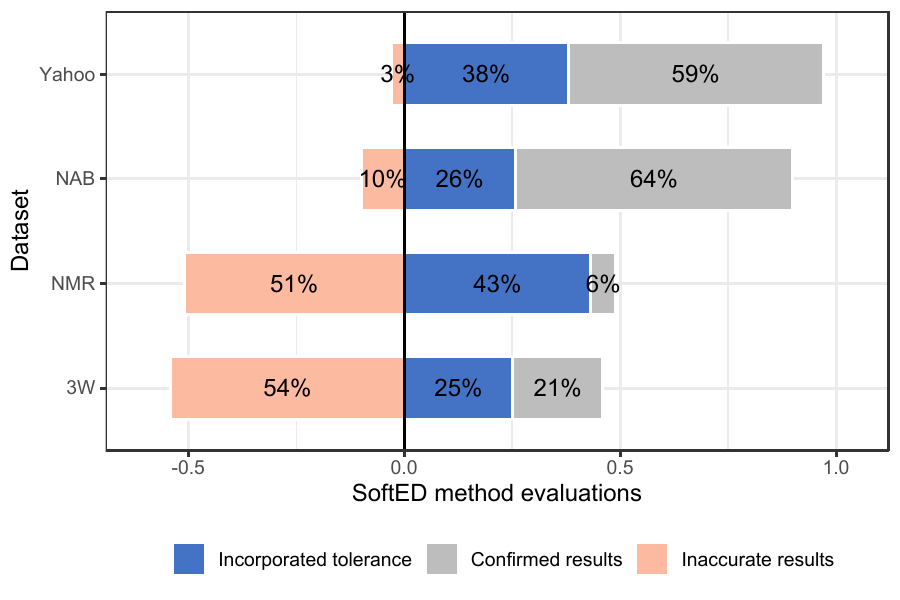}
 \caption{Incorporated temporal tolerance}
 \label{fig_exp11_1}
\end{subfigure}
\begin{subfigure}{.49\textwidth}
 \centering
 \includegraphics[width=1\linewidth]{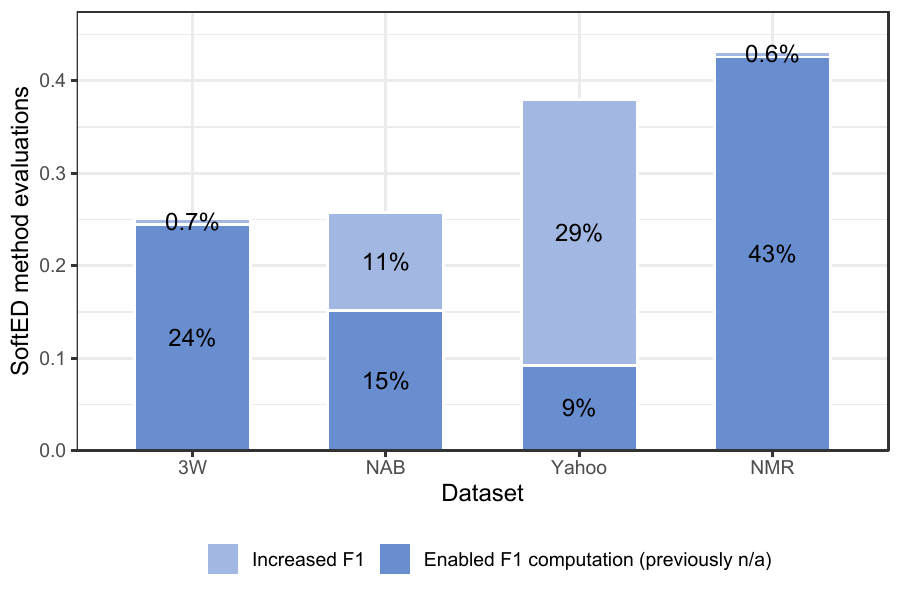}
 \caption{Detailed temporal tolerance opportunities}
 \label{fig_exp11_2}
\end{subfigure}
\caption{Incorporated temporal tolerance from SoftED F1 metric evaluation of event detectors compared to hard F1 metric.}
\label{fig_exp11}
\end{figure}

Figure \ref{fig_exp11_1} compares SoftED F1 and hard F1 metrics for each adopted dataset. The percentage of time series detections where SoftED F1 incorporated temporal tolerance is presented in blue. SoftED metrics incorporated temporal tolerance in over 43\% (NMR) and at least 25\% (3W) of detector evaluations in all datasets. In total, 36\% of the overall conducted time series detections were more tolerantly evaluated (in blue). Furthermore, 45\% of all detection results had their evaluation confirmed (in gray), maintaining an unchanged F1 score, reaching a maximum of 64\% for the NAB dataset and a minimum of 6\% for the NMR dataset. Finally, the other 19\% of the overall results corresponded to inaccurate detections that did not allow temporal tolerance (in red). The percentages of inaccurate results (F1 n/a) for each dataset are also in red in Figure \ref{fig_exp11_1}.

 Figure \ref{fig_exp11_2} details the cases of incorporated tolerance. The datasets NAB and Yahoo presented an increase in F1 in 11\% and 29\% of the cases, respectively (lighter blue). The other respective 15\% and 9\% are cases in which methods got no $TP$s, presenting zero Precision/Recall and non-applicable F1 based on hard metrics (darker blue). Nonetheless, SoftED could score sufficiently close detections, enabling the evaluation of such methods. This is also the case for almost all evaluations of the 3W and NMR datasets incorporating temporal tolerance. In fact, in total, 17\% of the overall conducted detection evaluations could not have been made without SoftED metrics incorporating temporal tolerance.

 It is possible to observe by Figure \ref{fig_exp11} that datasets that contained more anomalies (NAB and Yahoo) got more accurate detection results. It occurs as most adopted methods are designed for anomaly detection. Also, the number of anomaly events in their time series gives several opportunities for incorporating temporal tolerance and increasing F1. On the other hand, 3W and NMR datasets, containing only one or two change points per series, got a higher rate of inaccurate detections. These results indicate that change points pose a particular challenge for detection evaluation. SoftED metrics contribute by incorporating temporal tolerance whenever possible and scoring methods that could be disregarded.

\begin{figure}[ht!]
	\centering
	\includegraphics[width=0.49\textwidth]{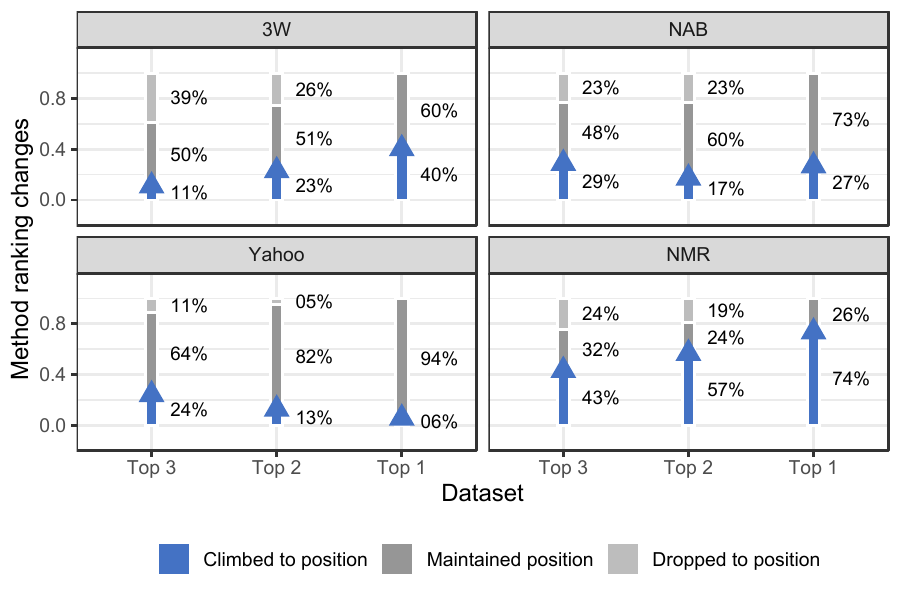}
	\caption{Changes in the ranking of top evaluated event detectors based on the SoftED F1 metric compared to hard F1 metric}\label{fig_exp12}
\end{figure}

\paragraph{Quantitative experiment 2} 
The second experiment focuses on whether the temporal tolerance incorporated by SoftED can affect the selection of different detectors. For that, we measured the number of times using SoftED metrics as criteria changed the ranking of the best-evaluated detectors. Figure \ref{fig_exp12} presents the changes in the top-ranked methods for each time series based on the SoftED F1 metric compared to the hard F1. For all datasets, there were changes in the best-evaluated detector (Top 1) in over 74\% (NMR) or at least 6\% (Yahoo) of the cases (in blue), affecting the recommendation of the most suitable detector for their time series. While the most accurate results maintained their top position (in dark gray), overall adopted time series, 31\% of detectors that could have been dismissed became the most prone to selection.

Furthermore, SoftED metrics also caused changes in the second (Top 2) and third-best (Top 3) evaluated methods. Percentages for each dataset are depicted in Figure \ref{fig_exp12}. Considering the adopted time series, 24\% of the methods in the Top 2 climbed to that position, while 16\% dropped to that position when other methods assumed the Top 1 (in light gray). For methods in the Top 3, 23\% climbed to the position, and in 24\% of the cases, they were pushed down by methods that climbed to the first two rank positions. Due to the higher rates of perfect recall results in the NAB and Yahoo datasets (Figure \ref{fig_exp11}), most of the methods applied maintained their ranking positions at the top. In contrast, 3W and NMR datasets presented more changes in ranking based on the SoftED metrics, affecting the selection of suitable methods, especially for change point detection.

\subsubsection{Complementary analysis} \label{quanti_discussion}

\begin{figure}[t!]
	\centering
	\includegraphics[width=0.49\textwidth]{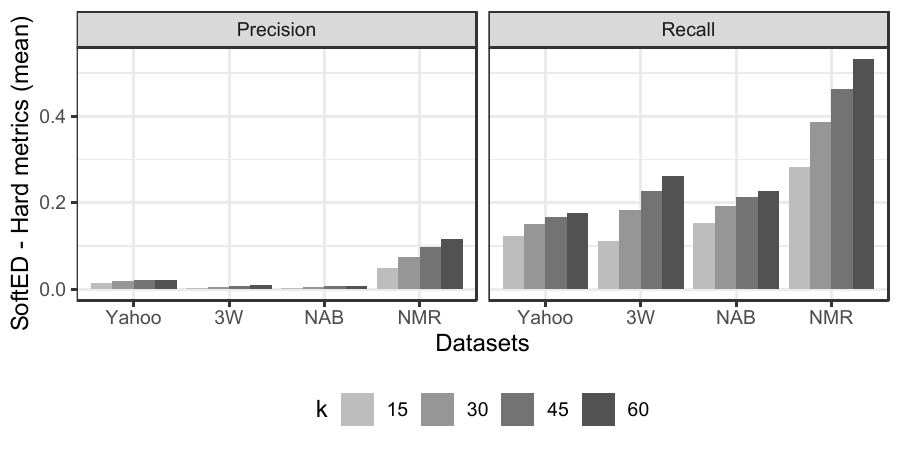}
	\caption{Average difference between SoftED and hard Precision and Recall metrics given different levels of temporal tolerance}\label{fig_exp13}
\end{figure}

Once the temporal tolerance and its effects in the ranking of detectors are, a complementary analysis can be performed to answer whether the definition of different levels of temporal tolerance affects the computation of the proposed metrics and to what extent. For that, one may conduct a sensitivity analysis by comparing the proposed metrics using different levels of temporal tolerance. The temporal tolerance level of SoftED metrics is given by the $k$ constant set to $30$, $45$, and $60$, besides the minimum value of $15$ as in the previous experiments. Figure \ref{fig_exp13} presents the average difference between SoftED and hard Precision and Recall metrics given the different levels of temporal tolerance for each dataset. As temporal tolerance increases, more $TP$s were considered, and metrics increased in value, which means the detectors were more tolerantly evaluated. In particular, higher levels of temporal tolerance lead to a decrease in the number of $FN$s, which most directly affected Recall values.

A final complementary experiment also aims to answer whether the temporal tolerance incorporated by the proposed metrics differs from that incorporated by NAB score anomaly windows. For that, we measured the number of times the NAB F1 metrics, derived from the NAB scoring algorithm, considered more $TP$s than hard F1 metrics while evaluating detectors. This measure is compared against the tolerance incorporated by SoftED metrics presented in Experiment 1 (Figure \ref{fig_exp11_1}). For datasets 3W, NAB, and Yahoo, NAB increased the incorporated tolerance at 42\%, 40\%, and 39\%, respectively. Whereas, for the NMR dataset, the percentage of incorporated tolerance decreased by 6\%.

NAB metrics were more tolerant than SoftED in method evaluations over most datasets, which does not mean better. The tolerance level incorporated by NAB depends directly on the anomaly window size, which is automatically set by the algorithm. Table \ref{tbl_exp14} presents the interval and the average of the anomaly window sizes set for the time series of each dataset. Since automatic anomaly window sizes set for NAB metrics change with each series, it did not allow a comparison under the same standardized conditions. While the tolerance level given by SoftED was consistently set by $k=15$, giving a tolerance window of $30$ observations, the NAB anomaly windows were mostly wider, reaching a maximum of $12,626$ observations or $1,357$ on average for the 3W dataset. Wider anomaly windows allow a greater number of hard $FP$s to be considered $TP$s, which causes F1 metrics to increase in value. It is similar to what was discussed in Experiment 3, explaining the increase in tolerance opportunities. The inverse is also true, as exemplified by the NMR dataset, for which anomaly windows did not surpass $14$ observations, decreasing the number of tolerance opportunities compared to SoftED.

\begin{table}[!ht]
\centering
\caption{Summary of anomaly window sizes automatically set by the NAB scoring algorithm for each dataset}
\label{tbl_exp14}
\setlength{\tabcolsep}{37pt}
\begin{tabular}{@{}ccc@{}}
\toprule
\multirow{2}{*}{\textbf{Dataset}} & \multicolumn{2}{l}{\textbf{Anomaly window sizes}} \\ \cmidrule(l){2-3} 
 & \textbf{Interval} & \textbf{Mean} \\ \midrule
3W & [52, 12626] & 1357 \\
NAB & [0, 902] & 286 \\
Yahoo & [0, 168] & 38 \\
NMR & [0, 14] & 13 \\ \bottomrule
\end{tabular}
\end{table}

It is also important to note from Table \ref{tbl_exp14} that the anomaly window size computation proposed by the NAB algorithm allows the definition of zero-sized windows, which do not give any tolerance to inaccuracy as in hard metrics or narrow windows, which are not enough to hold the central limit theorem guaranteed in SoftED results. On the other hand, the NAB window size automatic definition is not domain-dependent. Consequently, domain specialists may find windows too wide or too narrow for their detection application, making the incorporated tolerance and metric results non-applicable or difficult to interpret. In this context, SoftED contributes by allowing domain specialists to define the desired temporal tolerance level for their detector results.

\subsection{Qualitative analysis} \label{quali}

Although the analysis of Section \ref{quanti} is meant to evaluate the quantitative effects of temporal tolerance, it is not enough to answer whether these effects benefit the evaluation of event detection performance. Hence, there is a demand for a qualitative analysis of the contribution of the temporal tolerance incorporated by the proposed metrics under different scenarios. In this context, a given tolerant metric contributes if its adoption leads to selecting the ideal event detector for a given detection scenario. For that, we established a protocol (Section \ref{protocol}) and conducted the qualitative analysis (Section \ref{qualitative_results}).

\subsubsection{Metric evaluation protocol}\label{protocol}

This section proposes a general protocol for evaluating metrics in the context of time series event detection. In particular, this protocol is designed to evaluate metrics that incorporate temporal tolerance into detection. We established six competency questions (CQ)s \cite{uschold_ontologies_1996,noy_state_1997}, which encompass a set of questions stated and replied to in natural language \cite{noy_state_1997}. The proposed CQs for this evaluation protocol are designed to qualitatively analyze the contribution of temporal tolerance in event detection under different scenarios. A list of detection scenarios (one for each CQ) was established for that.

Each scenario corresponds to two different detectors (A and B) applied to a particular time series. Their detections are evaluated and compared based on the proposed metrics and other baseline and state-of-the-art metrics. The specialists can then assess the metrics that contribute most in any given scenario. Each scenario is associated with a specific CQ. To succeed in the evaluation, the proposed metric (SoftED) should be affirmative for all CQ.

The CQs (CQ1 to CQ6) are presented as follows. Each scenario (S1 to S6) characterizes the context for their evaluation.

\begin{itemize}
 \item[CQ1] Do the proposed metrics foster selecting the most suitable detector for precise detection (perfect recall)?

 \item[S1] Perfect recall: This scenario does not need temporal tolerance. All metrics should be able to reward perfect recalls.

 \item[CQ2] Do the proposed metrics foster selecting the most suitable detector approximate detection (partial recall)?

  \item[S2] Event neighborhood: Scenario in which methods produced detections that may not coincide with events but are in their surroundings (neighborhood). Tolerant metrics should be able to reward close detections.
  
 \item[CQ3] Do the proposed metrics foster selecting the most suitable detector for symmetric detections?

 \item[S3] Detection symmetry: Scenario in which methods produced symmetric detections. They have the same distance from the event and differ only in whether they come before or after it. Tolerant metrics should be able to reward close detections regardless of their relative position. 
 
 \item[CQ4] Do the proposed metrics foster selecting the most suitable detector with fewer false positives?

 \item[S4] Number of detections: Scenario in which methods produced detections close to the event contained in the series, differing only in the number of detections made. The closest detections have the same distance from the event. Tolerant metrics should be able to reward the closest detections while penalizing unnecessary $FP$s.
 
 \item[CQ5] Do the proposed metrics foster selecting the most suitable detector where detections are proximate to events?

 \item[S5] Detection distances: Scenario in which methods produced detections close to events differing only on their distance to them. Tolerant metrics should be able to reward the closest detections.
 
 \item[CQ6] Do the proposed metrics foster selecting the most suitable detector resilient of detection bias?

  \item[S6] Detection bias: Scenario in which metrics are prone to bias in detection evaluation. Methods produced both $TP$s and close detections before the event. Tolerant metrics should be able to reward close detections while maintaining the relative weight of $TP$s.
\end{itemize}

\begin{figure*}[!ht]
\centering
\begin{subfigure}{.40\textwidth}
 \centering
 \includegraphics[width=1\linewidth]{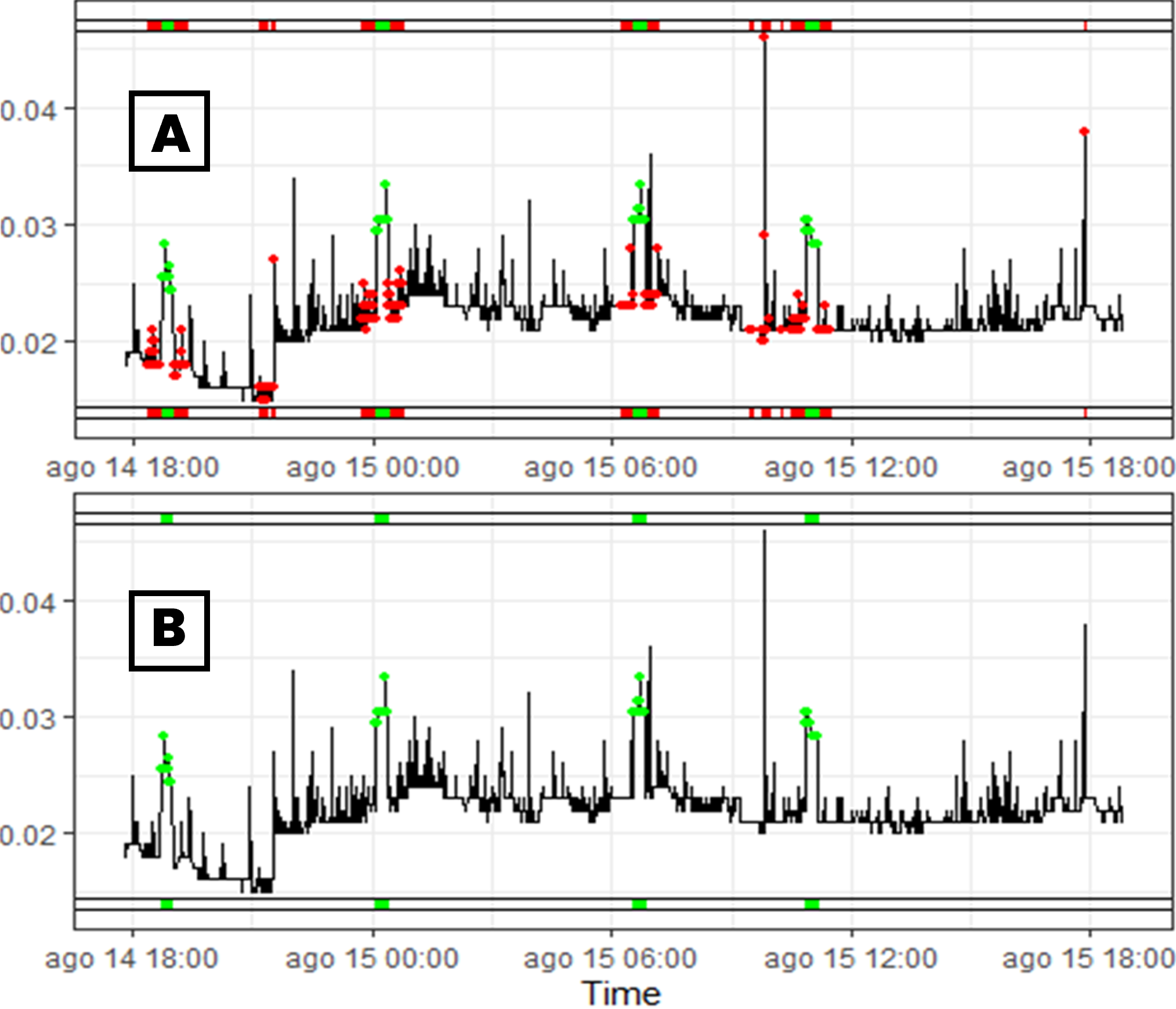}
 \caption{S1 - Perfect Recall}
 \label{fig_exp21}
\end{subfigure}
\begin{subfigure}{.40\textwidth}
 \centering
 \includegraphics[width=1\linewidth]{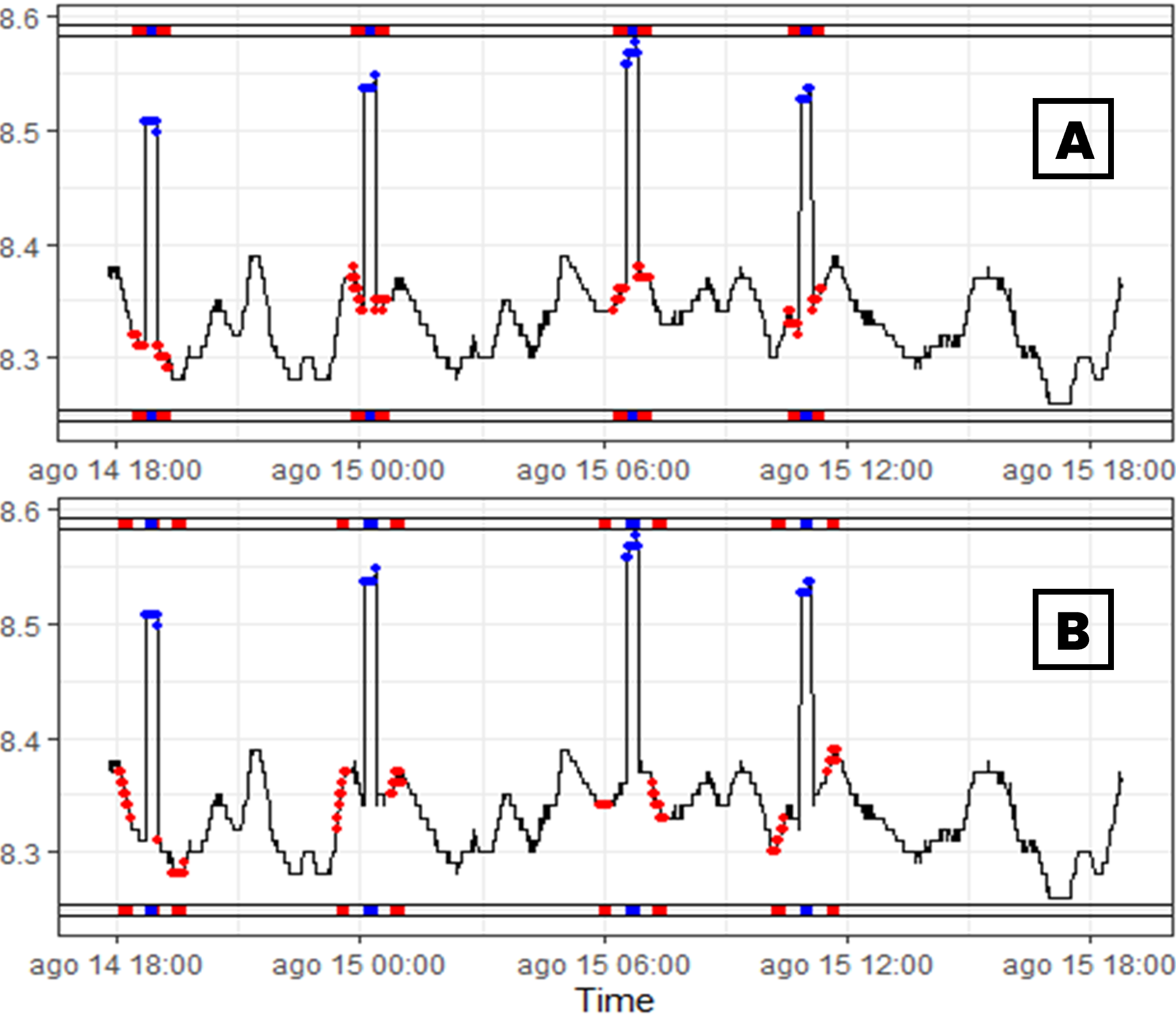}
 \caption{S2 - Event neighborhood}
 \label{fig_exp22}
\end{subfigure}
\begin{subfigure}{.40\textwidth}
 \centering
 \includegraphics[width=1\linewidth]{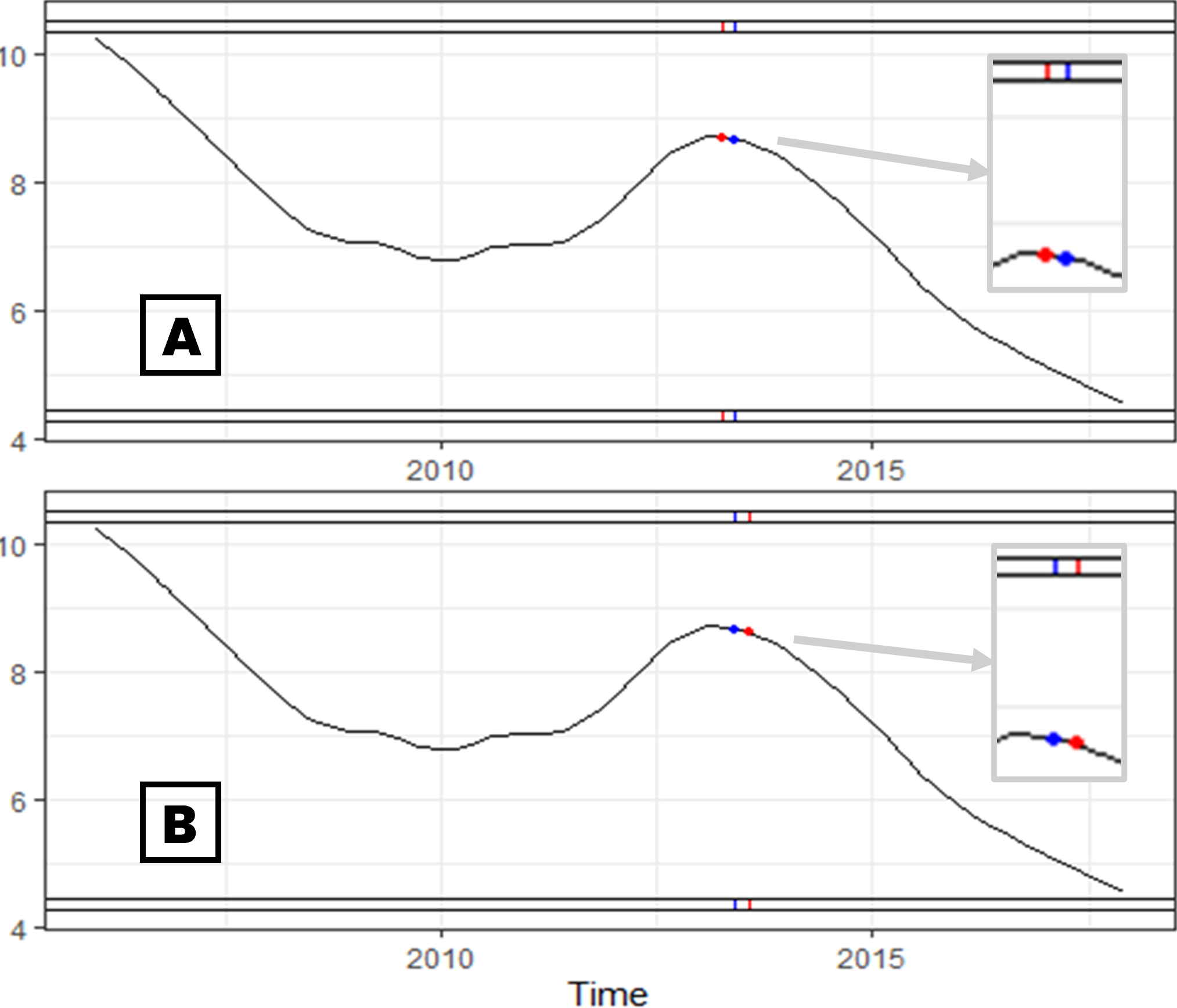}
 \caption{S3 - Detection symmetry}
 \label{fig_exp23}
\end{subfigure}
\begin{subfigure}{.40\textwidth}
 \centering
 \includegraphics[width=1\linewidth]{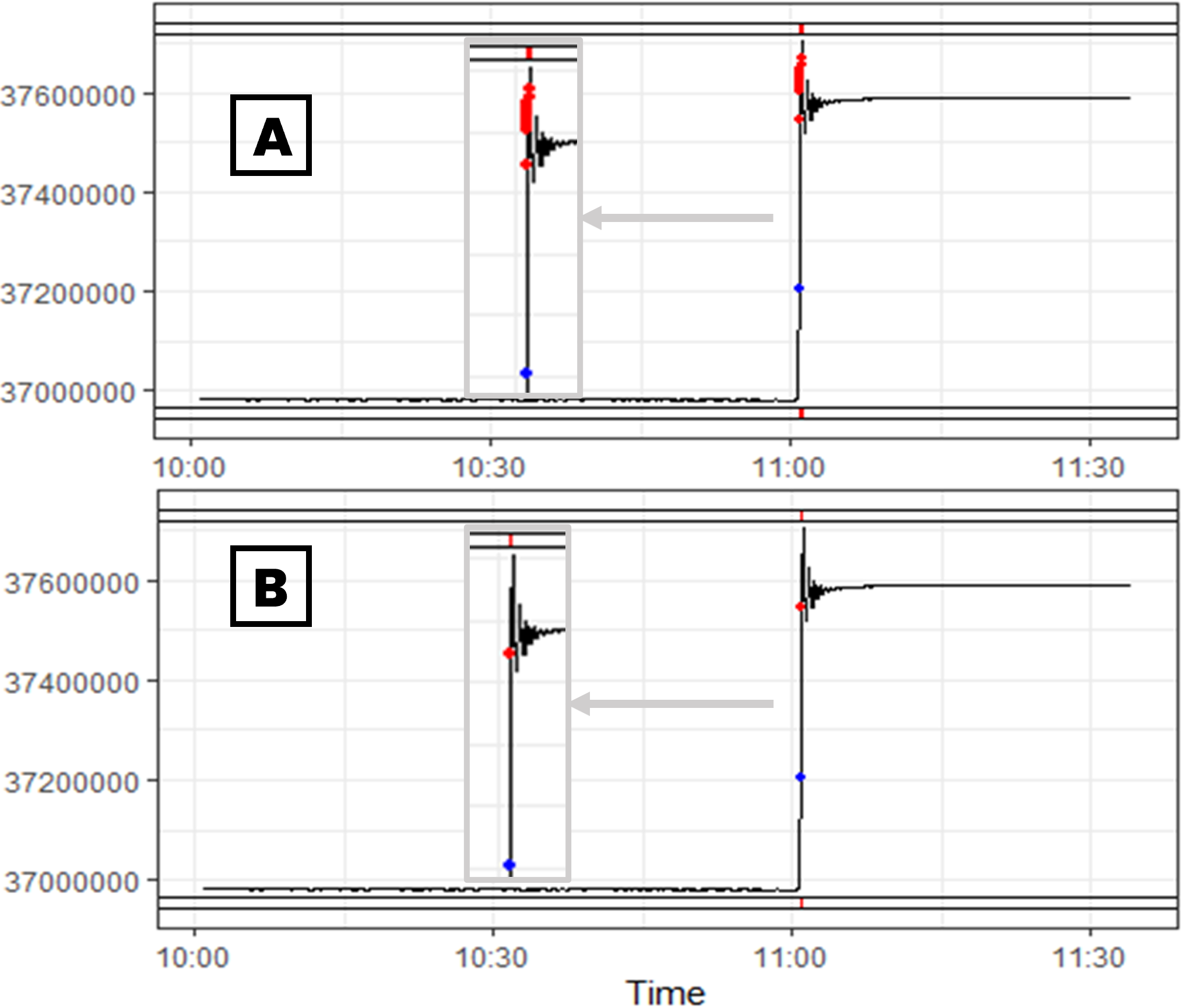}
 \caption{S4 - Number of detections}
 \label{fig_exp24}
\end{subfigure}
\begin{subfigure}{.40\textwidth}
 \centering
 \includegraphics[width=1\linewidth]{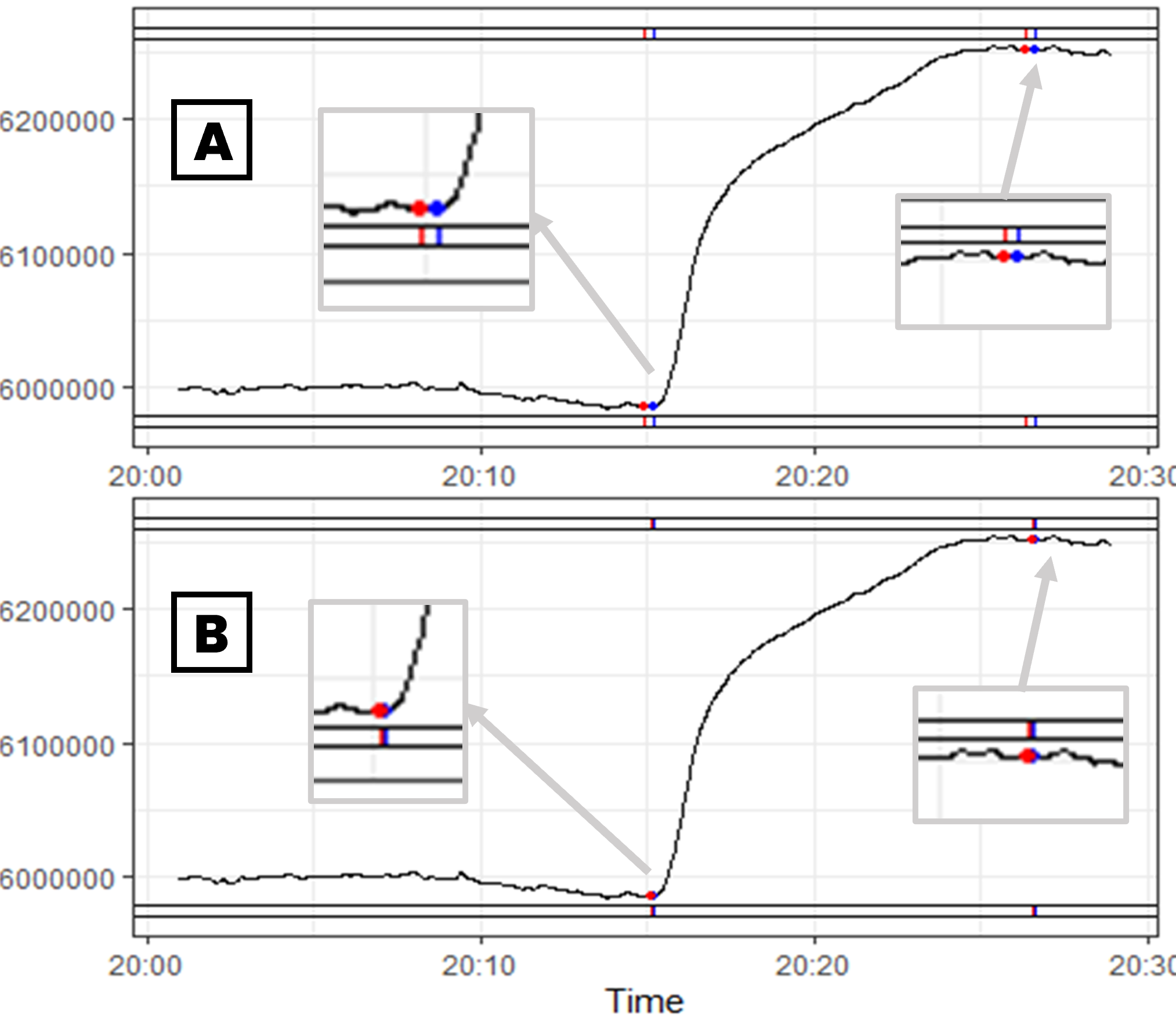}
 \caption{S5 - Detection distances}
 \label{fig_exp25}
\end{subfigure}
\begin{subfigure}{.40\textwidth}
 \centering
 \includegraphics[width=1\linewidth]{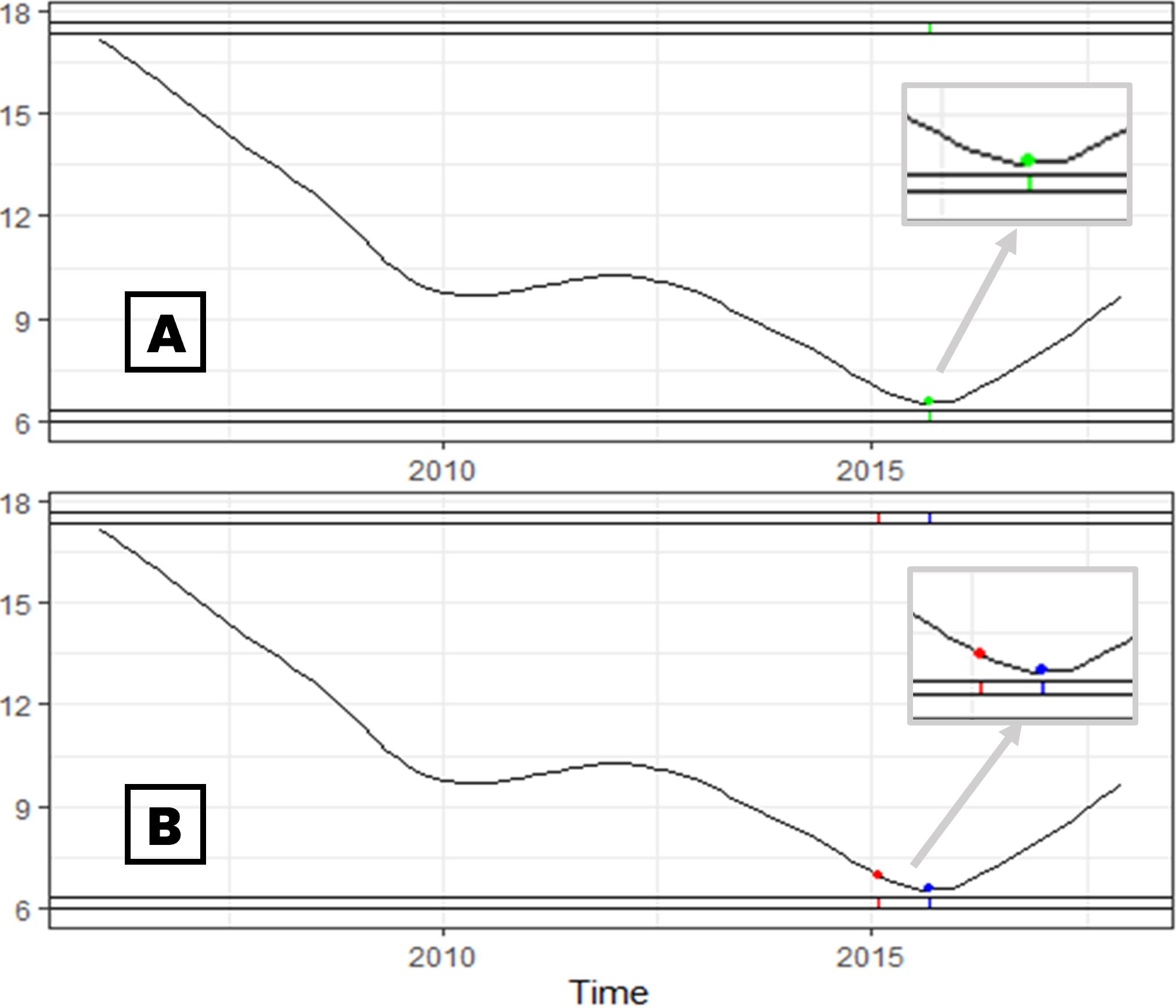}
 \caption{S6 - Detection bias}
 \label{fig_exp26}
\end{subfigure}
\caption{Detection results for the six scenarios comparing two given event detectors (A and B). Blue dots refer to time series events. Green and red dots refer to detections that coincide and do not coincide with events. (a) and (b) show the time series of the GECCO dataset (variables Trueb and pH). (c) and (f) show the time series of the NMR dataset (health facilities code 2080052 and 2295407). (d) and (e) show the time series of the 3W dataset (event type 2, variable P-PDG and event type 6, variable P-MON-CKP).}
\label{fig_exp2_plots}
\end{figure*}

The CQs are evaluated in a survey. The survey explores the problem of selecting the most suitable detector in six scenarios. Two detectors (A and B) were applied to each scenario's representative time series of the GECCO, 3W, or NMR datasets. The plots of the detection results were presented to participants as in Figure \ref{fig_exp2_plots}, where blue dots represent events, red dots represent detections, and green dots represent detections that match events. 

Table \ref{tbl_exp2_metrics} is also presented to all participants. It contains detection evaluation metrics computed for detectors A and B for each scenario, namely the F1 metric in its hard and SoftED versions and the NAB score. Values that maximize each metric and could be used to recommend a particular detector are underlined.

Given the results of both methods, the participants are requested to analyze the plots in Figure \ref{fig_exp2_plots} and answer the first question (Q1): 

\begin{itemize}
 \item[Q1] Which event detector performed better?
\end{itemize}

Q1 was closed with three disjoint options: \textit{Detector A}, \textit{Detector B}, or \textit{None}. The main goal of Q1 is to get the intuition and personal opinions of the most suitable detector for selection on that particular application scenario. 

Next, the participants are requested to analyze the metrics in Table \ref{tbl_exp2_metrics} and answer, for each scenario, the second question (Q2):

\begin{itemize}
 \item[Q2] Which metric corroborates with your opinion?
\end{itemize}

Q2 was also closed with three joint options: \textit{F1}, \textit{NAB score}, or \textit{Other}. The main goal of Q2 was to assess the metrics (and corresponding evaluation approach) that would foster the selection of the most suitable detector in that particular application scenario.

\subsubsection{Qualitative results}\label{qualitative_results}

This section presents a qualitative analysis of SoftED metrics compared to hard metrics and the NAB score. To this end, following the proposed metric evaluation protocol, we have surveyed $70$ participants, of which $13$ specialists from three domains: oil exploration, public health, and weather monitoring. We have interviewed three specialists from Petrobras (Brazil oil company), five specialists from the Oswaldo Cruz Foundation (Fiocruz), linked to the Brazilian Ministry of Health, the most prominent institution of science and technology applied to health in Latin America, and five weather forecast specialists from the Rio Operations Center (COR) of the City Hall of Rio de Janeiro. All interviewed specialists work on the problem of time series analysis and event detection daily as part of research projects. Furthermore, the other $57$ participants were volunteer students from the Federal Center for Technological Education of Rio de Janeiro (CEFET/RJ) and the National Laboratory for Scientific Computing (LNCC). They were introduced to characterize common sense interpretation.

\begin{table}[!ht]
\centering
\caption{Event detection metrics for detectors A and B for each scenario. Values that could be used to recommend a particular detector are underlined.}
\label{tbl_exp2_metrics}
\setlength{\tabcolsep}{8.5pt}
\begin{tabular}{@{}c c ccc@{}}
\toprule
\multirow{3}{*}{\textbf{Scenario}} & \multirow{3}{*}{\textbf{Detector}} & \multicolumn{3}{c}{\textbf{Metric}} \\ \cmidrule(l){3-5} 
 & & \multicolumn{2}{c}{\textbf{F1}} & \multirow{2}{*}{\textbf{NAB score}} \\
& & \textbf{Hard} & \textbf{SoftED} & \\ \midrule
\multirow{2}{*}{\textbf{S1}} & A & 0.40 & 0.40 & 16.05 \\
 & B & \underline{1} & \underline{1} & \underline{35.87} \\ \midrule
\multirow{2}{*}{\textbf{S2}} & A & n/a & \underline{0.07} & \underline{- 36.53} \\
 & B & n/a & 0.01 & - 50.17 \\ \midrule
\multirow{2}{*}{\textbf{S3}} & A & n/a & 0.87 & \underline{0.94} \\
 & B & n/a & 0.87 & 0.77 \\ \midrule
\multirow{2}{*}{\textbf{S4}} & A & n/a & 0.12 & 0.85 \\
 & B & n/a & \underline{0.6} & 0.85 \\ \midrule
\multirow{2}{*}{\textbf{S5}} & A & n/a & 0.07 & \underline{1.89} \\
& B & n/a & \underline{0.87} & 1.76 \\ \midrule
\multirow{2}{*}{\textbf{S6}} & A & 1 & \underline{1} & 0.88 \\
& B & n/a & 0.53 & \underline{1} \\ \bottomrule
\end{tabular}
\end{table}

Table \ref{tbl_exp2_responses} presents the domain specialists' responses to the questions for each scenario. Their winning responses are underlined. Furthermore, student volunteers' winning responses are given to study how specialist opinion compares with common sense. All participants were also allowed to comment and elaborate on their responses for each scenario in an open question. The remainder of this section further discusses the results of each scenario. 

\begin{table*}[!ht]
\centering
\caption{Specialists' responses to the questions for each scenario. The winning responses are underlined. Volunteers winning responses are also given for comparison with the non-specialist common sense.}
\label{tbl_exp2_responses}
\setlength{\tabcolsep}{6.4pt}
\begin{tabular}{@{}c cccccc cc@{}}
\toprule
\multirow{3}{*}{\textbf{Scenario}} & \multicolumn{6}{c}{\textbf{Specialists responses}} & \multicolumn{2}{c}{\textbf{Volunteer winning responses}} \\ \cmidrule(l){2-9} 
 & \multicolumn{3}{c}{\textbf{Q1}} & \multicolumn{3}{c}{\textbf{Q2}} & \multirow{2}{*}{\textbf{Q1}} & \multirow{2}{*}{\textbf{Q2}} \\ \cmidrule(lr){2-7}
 & \textbf{Detector A} & \textbf{Detector B} & \multicolumn{1}{c}{\textbf{None}} & \textbf{F1} & \textbf{NAB score} & \multicolumn{1}{l}{\textbf{Other}} & & \\ \midrule
\textbf{S1} & 1 (8\%) & \underline{12 (92\%)} & \multicolumn{1}{c}{0 (0\%)} & \underline{12 (92\%)} & 6 (46\%) & 1 (8\%) & Detector B (84\%) & F1 (96\%) \\
\textbf{S2} & \underline{11 (84\%)} & 1 (8\%) & \multicolumn{1}{c}{1 (8\%)} & \underline{12 (92\%)} & 7 (54\%) & 1 (8\%) & Detector A (88\%) & F1 (96\%) \\
\textbf{S3} & \underline{12 (92\%)} & 0 (0\%) & \multicolumn{1}{c}{1 (8\%)} & 4 (31\%) & \underline{12 (92\%)} & 0 (0\%) & Detector A (84\%) & NAB score (82\%) \\
\textbf{S4} & 0 (0\%) & \underline{12 (92\%)} & \multicolumn{1}{c}{1 (8\%)} & \underline{13 (100\%)} & 1 (8\%) & 0 (0\%) & Detector B (86\%) & F1 (98\%) \\
\textbf{S5} & 3 (23\%) & \underline{10 (77\%)} & \multicolumn{1}{c}{0 (0\%)} & \underline{11 (85\%)} & 5 (38\%) & 0 (0\%) & Detector B (74\%) & F1 (86\%) \\
\textbf{S6} & \underline{10 (77\%)} & 3 (23\%) & \multicolumn{1}{c}{0 (0\%)} & \underline{10 (77\%)} & 6 (46\%) & 0 (0\%) & Detector A (82\%) & F1 (89\%) \\ \bottomrule
\end{tabular}
\end{table*}

\paragraph{CQ 1} The first scenario refers to a scenario of perfect recall, where Detector A and Detector B detected all events in the GECCO dataset. However, Detector A presents more details (in red). Table \ref{tbl_exp2_responses} shows that almost all specialists (12/13) agreed that Detector B performed better. According to them, Detector B managed to minimize $FP$s, presenting a higher Precision rate, indicated by the F1 metric, which was also the winning response for Q2 with 12/13 votes. For this scenario, both hard and SoftED F1 give the same evaluation of Detector B so that both approaches can be used for recommendation.

Nonetheless, $6$ specialists (46\%) also selected the NAB score, which also corroborates with the recommendation of Detector B. Other specialists said they preferred not to select the NAB score, as they were unfamiliar with the metric and wanted to avoid concluding with this scenario. 

\paragraph{CQ 2} The second scenario is based on another time series from the GECCO dataset. It addresses the scenario in which Detector A and Detector B presented detections that, despite not coinciding with the events contained in the series, are in the surroundings or the neighborhood of the events. Furthermore, Detector A and Detector B detections differ in the distance to events. In this case, most specialists (11/13) agreed to select Detector A as giving the best detection performance, for their detections are temporally closer to the events. Both metrics, F1 and NAB score, corroborated with specialists' opinions recommending Detector A, while F1 was the winning response to Q2. At this point, it is important to note that the hard approach to F1 computation can no longer evaluate the methods, as both results had no Precision or Recall. Hence, the winning response for Q2 regards the F1 metric produced by the SoftED approach as the one that fosters the selection of the best detection performance according to specialists.

\paragraph{CQ 3} The third scenario is based on a time series from the NMR dataset containing monthly neonatal mortality rates for a healthcare facility in Brazil over the years. In this scenario, Detector A and Detector B produced only one detection close to the event contained in the series. The detections of Detector A and Detector B are symmetric. They have the same distance from the event and differ only in whether they come before or after it. In this scenario, almost all specialists (12/13) responded that Detector A gave the best detection performance, as it seems to anticipate the event, allowing time to take prior needed actions. Furthermore, as there was a tie regarding the F1 metrics, the NAB score was the winning response, corroborating with specialists' opinions. 

However, a public health specialist from Fiocruz disagreed and responded that none of the methods performed better, which is corroborated by the F1 metrics. For example, consider implementing a public health policy in which a human milk bank is supposed to decrease neonatal mortality rates. Although it makes sense to detect the first effects of preparing for the implementation of the policy, it may not be reasonable to give greater weight to anticipated detections rather than the detection of the effects after the implementation. They defend:
\begin{quote}
It is important to deepen the understanding of the context of the event and the reach of its effects (before and after).
\end{quote}

\paragraph{CQ 4} The fourth scenario is based on a time series taken from the 3W dataset produced by Petrobras. In this scenario, Detector A and Detector B presented detections close to the event contained in the series, differing only in the number of detections made. The closest detections for both methods have the same distance from the event. For this scenario, except for one specialist that responded \textit{None} to Q1, all specialists agree that Detector B performed better. As it minimizes the overall $FP$s, it increases Precision, which conditions F1, the winning response of Q2, selected by 100\% of the specialists. The NAB score indicates a tie between both methods, therefore not penalizing the excess $FP$s, and the hard F1 does not provide any evaluation. In this case, the SoftED F1 metric is the only one corroborating with specialists' opinions.

\paragraph{CQ 5} The fifth scenario is based on another time series from the 3W dataset. This scenario addresses the problem of evaluating methods based on their detection proximity to events. In this scenario, both Detector A and Detector B presented a detection close and antecedent to the two events contained in the series. Detector A and Detector B differ only concerning the distance of their detections to the events. Most specialists (10/13) agreed that Detector B performed better, as they say:
\begin{quote}
Giving greater weight to detections closer to the actual events seems reasonable.
\end{quote}
Again, the only metric that corroborated the specialists' opinion was the SoftED F1. Furthermore, specialists mentioned that SoftED F1 was approximately 12 times greater for Detector B than Detector A, while the difference in the NAB score did not seem high enough to give the same confidence in Detector A's results.

\paragraph{CQ 6} Finally, we used another time series of neonatal mortality rates from the NMR dataset for the sixth and final scenario. This scenario addresses the problem of detection bias in detection evaluation. Detector A and Detector B produced a detection related to the event contained in the series. However, Detector A and Detector B detections differ regarding their distance to the event. Detector A managed to correctly detect the time series event, while the detection of Detector B came close before the event. Most specialists (10/13) agreed that Detector A performed better since it produced, for all intents and purposes, a $TP$, presenting perfect recall and perfect Precision. On the other hand, the evaluation of Detector B depended solely on the incorporation of temporal tolerance.

As metrics disagree with the recommendation, the F1 metric again corroborates with the specialists' opinion, being the winning response for Q2. In particular, the SoftED F1 metric is the only approach that recommends Detector A. The hard F1 metric cannot be computed for Detector B, being incomparable. The difference in metric values of SoftED is also greater than for the NAB score, increasing confidence in the recommendation.

\subsubsection{Discussion of results} \label{quali_discussion}

Given different detection evaluation scenarios, the majority of the domain specialists agreed that the most desired detector for selection was the one that minimizes $FP$s and $FN$s, giving higher Precision and Recall rates while also producing detections that are temporally closer to the events. In this context, the F1 was the metric most corroborated with specialists' opinions for $5$ of the $6$ scenarios. In particular, according to specialists in four scenarios, the SoftED F1 metric was the only one that furthered the selection of the most desired detector. To elaborate, a domain specialist from Fiocruz argued that:
\begin{quote}
For health policies, for example, the SoftED approach makes more sense since the hard and NAB approaches do not seem adequate for events that produce prior and subsequent effects that may have a gradual and non-monotonous evolution.
\end{quote}
Volunteer winning responses in Table \ref{tbl_exp2_responses} also indicate that common sense does not differ from specialist opinion, which means the contribution of SoftED metrics is noticeable even to a wider non-specialist public.

There was still one scenario (S3) where the NAB score was the metric that most corroborated with specialists. They claimed that for their usual detection application, it is interesting to have $FP$s (warnings) before the event, so there is a time window for measures to be taken to prevent any of its unwelcome effects. Also, the longer the time window set by $FP$s preceding the event, the better, as there is more valuable time to take preventive actions. 

It is important to mention that, to avoid bias in the responses, the discussion regarding our motivating example of Section \ref{motivating_example} was not presented before the interviews. Hence, detections that preceded the events were misconceived as event predictions \cite{zhao_event_2021}. As presented earlier, detections preceding events can be made past their occurrence. Evaluating methods that anticipate events is not about how temporally distant a preceding detection is from the events. It is actually about the time lag needed for a detector to detect the event accurately. The misconception regarding detections that preceded the events was addressed in detail by the end of the interviews. Furthermore, the specialists see the evaluation regarding detection lags, that is, analyzing its ability to anticipate (or not) the events as complementary to detection performance but also important. 

\section{Conclusions}\label{conclusion}

This paper introduced the SoftED metrics, which are new softened versions of the standard classification metrics. Inspired by Fuzzy sets, they are designed to incorporate temporal tolerance in evaluating event detection performance in time series applications. SoftED metrics support the comparative analysis of methods based on their ability to accurately produce detections of interest to the user, given their desired tolerance level.

This paper also introduces a new general protocol inspired by competency questions to evaluate temporal tolerant metrics for event detection. Following the proposed evaluation protocol, the SoftED metrics were quantitatively and qualitatively compared against the current state-of-the-art methods. The metrics incorporated temporal tolerance in event detection, enabling evaluations that could not have been made without them while also confirming accurate results. Moreover, specialists noted the contribution of SoftED metrics to the problem of detector evaluation in different domains. 
	
	\section*{Acknowledgments}
	The authors thank CNPq, CAPES (finance code 001), FAPERJ, and CEFET/RJ for partially funding this research.
	
	\section*{Conflict of interest}
	On behalf of all authors, the corresponding author states that there is no conflict of interest.

	\bibliographystyle{abbrv}
	
\end{document}